\documentclass[runningheads]{llncs}

\usepackage[preprint]{eccv}

\usepackage{eccvabbrv}

\usepackage{graphicx}
\usepackage{booktabs}
\usepackage{multirow}
\usepackage[table]{xcolor}
\usepackage{tcolorbox}
\usepackage{wrapfig}
\usepackage[accsupp]{axessibility}  %

\usepackage{hyperref}
\usepackage{xr-hyper}
\makeatletter
\def\blfootnote{\gdef\@thefnmark{}\@footnotetext}
\makeatother
\begin{document}

\title{Learning from Reliable Negatives: Confidence-Anchored Test-Time Adaptation for GUI Grounding} 

\titlerunning{Confidence-Anchored TTA for GUI Grounding}

\author{Yizhou Liu\inst{*1} \and Fei Tang\inst{*1} \and Yuchen Yan\inst{1} \and Zhengxi Lu\inst{1} \and Songqin Nong\inst{2} \and \\ Tao Jiang\inst{2} \and Wenhao Xu\inst{2} \and Wenqi Zhang\inst{1} \and Weiming Lu\inst{1} \and Jun Xiao\inst{1} \and \\Yongliang Shen\inst{\dagger1}}

\authorrunning{Y.~Liu et al.}

\institute{Zhejiang University \and Ant Group\\
\email{\{22551294,syl\}@zju.edu.cn}}

\maketitle
\blfootnote{$^*$ Equal contribution, $^\dagger$ Corresponding author.}

\begin{abstract}
  Graphical User Interface (GUI) grounding is essential for autonomous agents to map natural language instructions to precise screen coordinates. However, existing supervised fine-tuning and reinforcement learning methods are constrained by the high cost of annotation, creating a scalability bottleneck. In this paper, we introduce a label-free test-time training paradigm driven by two key insights: (1) confidence patterns in coordinate tokens are a better indicator than full-sequence confidence, and (2) in sparse GUI coordinate spaces, negative samples offer more reliable learning signals than potentially noisy positive ones. We first propose \textbf{C}onfidence-\textbf{A}nchored \textbf{L}earning (CAL), which utilizes coordinate-token confidence to filter pseudo-labels and assign distance-based binary rewards. Building on this, we develop \textbf{C}onfidence-\textbf{A}nchored \textbf{N}egative \textbf{L}earning (CANL), which exclusively optimizes the model using negative samples to bypass the risks of incorrect positive samples. Experimental results demonstrate that CANL-7B achieves 92.1\% on ScreenSpot-V2. On more challenging ScreenSpot-Pro, CANL-7B reaches 33.8\%, an 8.9\% absolute improvement over the base model. Our findings establish coordinate-token confidence as a powerful alternative to manual annotations for scalable GUI agent development. 
\end{abstract}

\section{Introduction}
GUI agents have emerged as a critical technology for automating human-computer interaction, enabling natural language commands to be translated into precise interface actions. While early rule-based systems required extensive manual engineering and lacked adaptability \cite{zhou2024webarenarealisticwebenvironment, mobileagent}, the advent of multimodal large language models (MLLMs) \cite{qwen-vl, gpt4o, qwen2.5-vl, qwen3-vl ,zhou2024learningobservergazezeroshotattention,bee} has opened new possibilities for flexible GUI understanding and interaction \cite{tang2025survey, osagent, mai-ui, mobilerl, step-gui, ui-venus, mobile_agent_v3.5, gui-shift, phi-ground,clawgui}. At the heart of these agents lies GUI grounding, the task of mapping natural language instructions to specific interface elements through coordinate prediction \cite{seeclick, showui, osatlas}.

\begin{figure}
    \centering
    \includegraphics[width=0.98\linewidth]{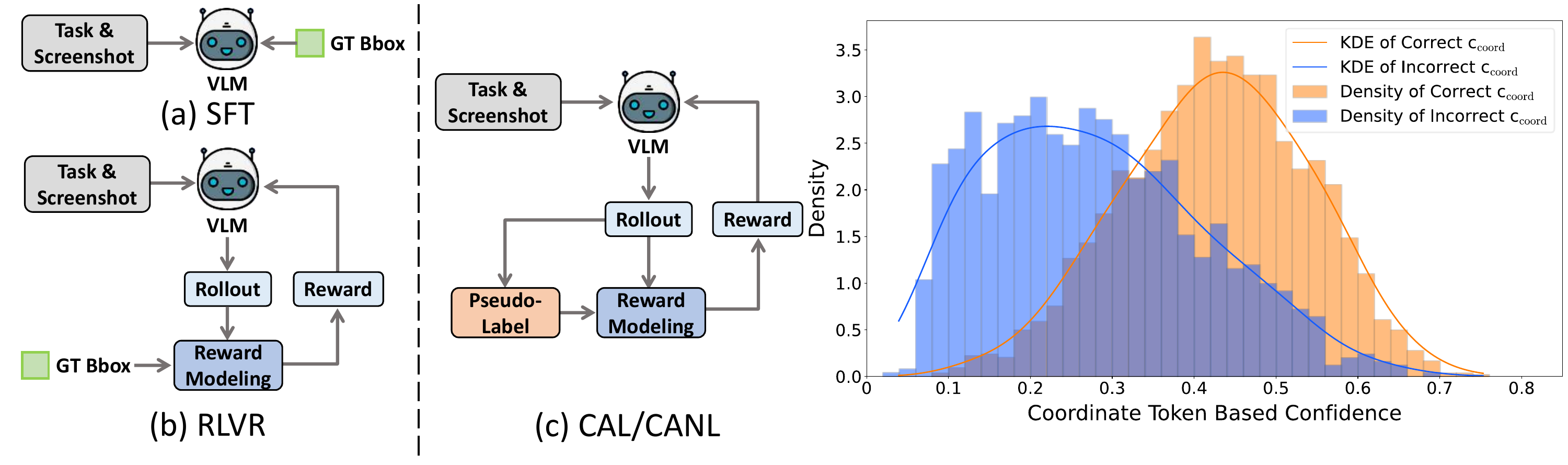}
    \caption{Training paradigms for GUI grounding and validation of coordinate-token confidence. \textbf{Left}: Comparison of training methods: (a) SFT and (b) RLVR require ground truth labels, while (c) our CAL/CANL methods leverage this confidence signal for label-free training. \textbf{Right}: Distribution of coordinate-token confidence on ScreenSpot-V2 shows clear separation between correct and incorrect predictions, with correct predictions exhibiting significantly higher confidence values. }
    \label{fig:intro_ans}
\end{figure}

Current approaches to GUI grounding employ two main training paradigms, as illustrated in \cref{fig:intro_ans}: (1) Supervised Fine-Tuning (SFT) \cite{cogagent, seeclick, uground, osatlas, showui, focus, aguvis, ui-tars, gui-actor, sun2025osgenesisautomatingguiagent}, which directly learns from ground-truth bounding box annotations, and (2) Reinforcement Learning with Verifiable Rewards (RLVR) \cite{ui-r1, gui-r1, infigui-r1, gui-g1, gui-se, lpo, infigui-g1, ui-s1, guirlvg, points-gui-g, ui-ins}, which requires ground-truth labels to compute accurate rewards for policy optimization. However, both paradigms face a fundamental challenge: the heavy reliance on annotated data. Obtaining pixel-level bounding box annotations is prohibitively expensive and time-consuming, requiring manual labeling of precise coordinates for each UI element. This dependency on labeled data severely limits the scalability and practical deployment of GUI agents across diverse applications and domains.

A natural question arises: can we enhance GUI grounding capabilities without explicit supervision? Test-time reinforcement learning (TTRL) \cite{ttrl} offers a promising direction, having demonstrated success in mathematical reasoning tasks through label-free adaptation. However, applying TTRL to GUI grounding presents unique challenges. The region consistency supervision based on predicted bounding boxes introduces large supervision errors, leading to marginal performance gains \cite{gui-rcpo}. Unlike mathematical problems where correctness can be verified through symbolic computation, GUI tasks require spatial reasoning over visual elements where ground truth is unavailable during inference.

To address this challenge, we propose leveraging the model's own confidence signals as a source of supervision. Our key insight is that while model's most confident predictions may not always be correct, its coordinate-level token probabilities provide valuable signals for distinguishing likely correct from incorrect predictions. As illustrated in \cref{fig:intro_ans}, we empirically observe that the confidence distributions of correct and incorrect predictions form distinct, well-separated clusters when focusing on coordinate tokens. This observation motivates our development of coordinate-token confidence, a metric that focuses specifically on the probability values of tokens representing spatial coordinates rather than the entire response sequence.

Building on this foundation, we introduce \textbf{C}onfidence-\textbf{A}nchored \textbf{L}earning (CAL), which generates multiple candidate predictions, selects the most confident one as a pseudo-label based on coordinate-token confidence, and performs reinforcement learning using binary rewards derived from spatial distances to this pseudo-label. As shown in \cref{fig:intro_ans}(c), our approach eliminates the need for ground truth labels by using these pseudo-labels to assign rewards directly, enabling truly label-free training.
However, pseudo-labels inherently contain errors that could negatively impact learning. This leads to a critical observation: in the sparse coordinate space of GUI grounding, negative samples (predictions far from the pseudo-label) are overwhelmingly likely to be incorrect, while positive samples near potentially misplaced pseudo-labels may be unreliable. This asymmetry motivates \textbf{C}onfidence-\textbf{A}nchored \textbf{N}egative \textbf{L}earning (CANL), 
which modifies the advantage computation to zero out potentially unreliable positive samples while preserving negative learning signals. By focusing solely on reliably incorrect predictions, CANL transforms the challenge of pseudo-label uncertainty into an opportunity for robust learning.

Evaluation across four benchmarks validates our approach. Without any annotations, CANL-7B achieves 92.1\% on ScreenSpot-V2, bringing a 4\% performance improvement. CANL-7B reaches 33.8\% on ScreenSpot-Pro, an 8.9\% absolute improvement over the base model, exceeding GUI-R1-7B which requires ground truth rewards. On challenging benchmarks ScreenSpot-Pro and UI-Vision, CANL consistently outperforms CAL by 0.6-1.5\%, confirming that negative samples provide more reliable supervision when targets are small and sparse. Analysis reveals coordinate-token confidence outperforms eight alternative pseudo-labeling strategies by 2.1-11.9\%, while CANL maintains stable performance across predefined distance thresholds compared to CAL's variance, demonstrating superior robustness for practical deployment. Furthermore, Experiments on AndroidWorld demonstrate that the grounding performance gains delivered by our methods can be effectively applied to GUI navigation tasks.

Our contributions can be summarized as follows:
\begin{itemize}

\item We propose CAL, a label-free training approach for GUI grounding that introduces coordinate-token confidence to identify optimal predictions among multiple candidates and uses these as pseudo-labels for reinforcement learning without manual annotations.
\item We further develop CANL, which exclusively uses negative samples during reinforcement learning, demonstrating that learning from reliably incorrect predictions can be more effective than using potentially inaccurate positive samples in label-free settings.
\item Extensive experiments demonstrate that our label-free approaches achieve competitive performance across GUI Grounding benchmarks.

\end{itemize}

\section{Related Work}

\subsection{GUI Grounding}
GUI grounding bridges natural language instructions with GUI elements, enabling agents to understand and interact with software environments through grounded multimodal reasoning \cite{seeclick,uisage,uizoomer}. Given a screenshot and a natural language command, the task requires identifying the corresponding UI element and returning its location as either a bounding box or a point coordinate. Early efforts in enhancing GUI grounding capabilities primarily relied on supervised fine-tuning (SFT) \cite{seeclick, uground, cogagent, osatlas, aguvis} using large-scale annotated datasets. Building on the success of GRPO and DeepSeek-R1 \cite{grpo, r1}, recent work has shifted toward RLVR, where models leverage interaction signals and feedback to further enhance grounding performance \cite{ui-r1, gui-r1, infigui-r1, gui-g1, gui-se, guig2, guirlvg, gta1}. However, both SFT and RLVR paradigms fundamentally depend on labeled data, where manual annotation is prohibitively expensive and automated labeling processes often introduce errors, creating a major bottleneck for scaling GUI agents to diverse applications and domains. GUI-RCPO \cite{gui-rcpo} adopts unsupervised training with predicted bounding boxes, yet erroneous supervision yields marginal gains.

\subsection{Negative Learning}
Negative learning (NL) reframes supervised learning by training models on what to avoid rather than what to produce. It operates on the principle that model's least confident predictions serve as reliable negative signals, even when its most confident predictions may be incorrect \cite{negative_learning}. This approach has proven effective for tasks with noisy labels and in few-shot settings where robust training signals are scarce \cite{few-shot_nl}. Recently, NL has been applied to large language models for mathematical reasoning, using either an increased ratio of negative samples \cite{bridging_sl_rl, learning_from_failue} or negative samples exclusively \cite{nl_for_math}. Despite its success in text-based domains, the application of NL to multimodal models remains unexplored. Our work explore whether learning solely from negative examples can be an effective strategy for complex vision-language tasks such as GUI grounding.
\begin{figure}[t]
    \centering
    \includegraphics[width=0.95\linewidth]{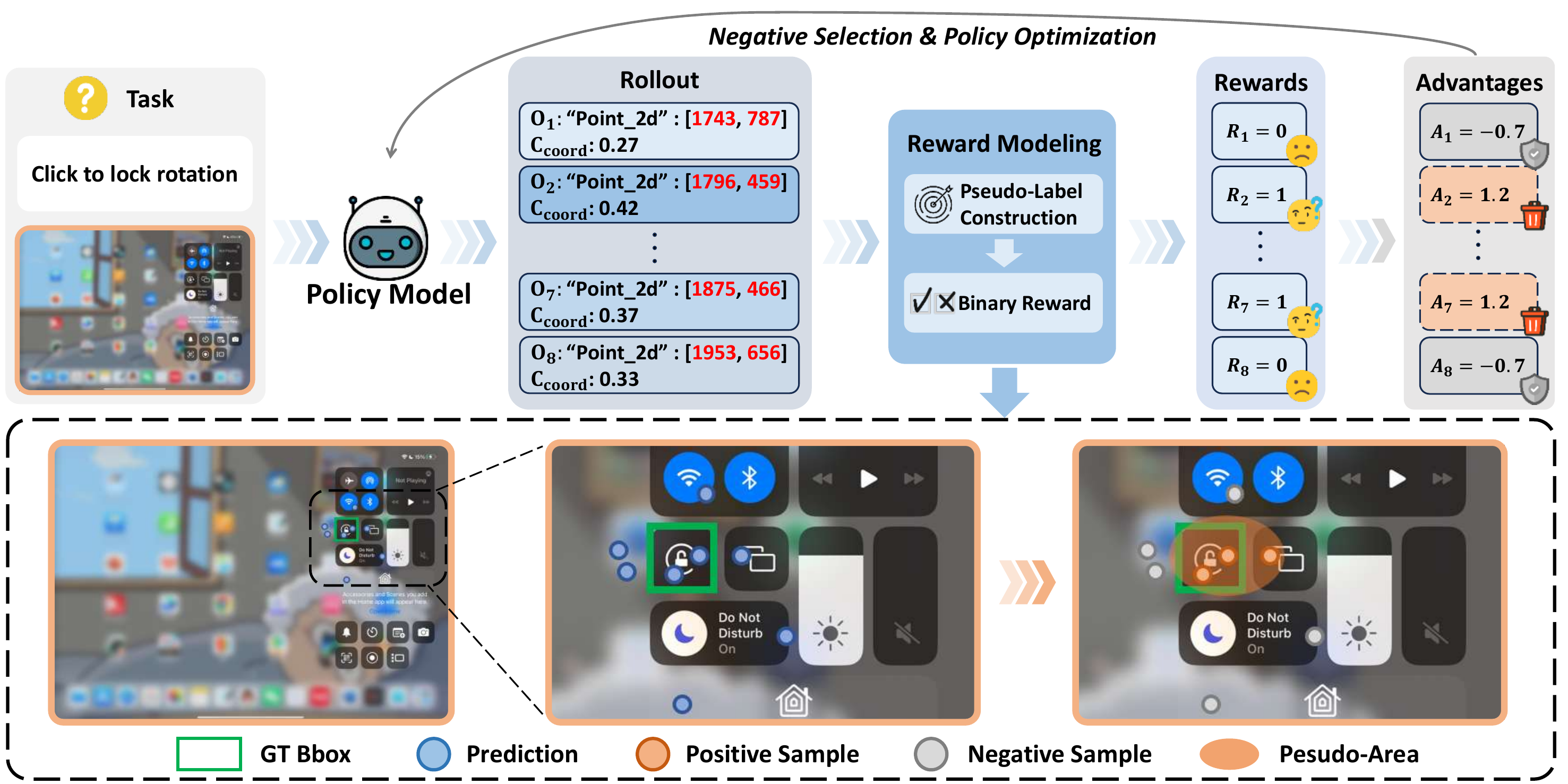}
    \caption{The CANL pipeline. Given an instruction and screenshot, the model generates multiple predictions with associated coordinate-token confidence scores, selects the highest-confidence prediction as a pseudo-label, constructs a pseudo-area using the distance threshold $\tau$, and assigns binary rewards. During policy optimization, CANL leverages only negative samples (gray) while zeroing advantages for potentially unreliable positive samples (orange), as illustrated in the bottom panel showing the transformation from standard to negative-only reinforcement learning.}
    
    \label{fig:method}
\end{figure}

\section{Method}
\label{sec: method}
We propose a label-free training paradigm for GUI grounding. Our approach consists of two key components: (1) coordinate-token confidence-based pseudo-label generation that selects the most reliable prediction from multiple samples, and (2) distance-based reinforcement learning that assigns rewards without ground truth. We present two variants: \textbf{C}onfidence-\textbf{A}nchored \textbf{L}earning (CAL) that performs policy optimization using both positive and negative rewards, and \textbf{C}onfidence-\textbf{A}nchored \textbf{N}egative \textbf{L}earning (CANL) that exclusively learns from negative samples to mitigate pseudo-label errors in the GUI coordinate space.

\subsection{Confidence-Based Pseudo-Label Generation}

The core challenge in label-free GUI grounding is identifying reliable training signals from the model's own predictions. Standard confidence estimation averages probabilities across all tokens:
\begin{equation}
\label{confidence}
c(y \mid x) = \frac{1}{n} \sum_{i=1}^{n} P(y_i \mid y_1, \dots, y_{i-1}, x)
\end{equation}
However, this approach fails in GUI tasks as high-probability formatting and descriptive tokens mask the uncertainty inherent in critical coordinate values. Our analysis reveals that coordinate tokens—the numerical values specifying pixel locations—carry the primary uncertainty signal, with 71.5\% having probabilities below 0.6 while 78.1\% of non-coordinate tokens exceed 0.9 on ScreenSpot-V2. We therefore introduce coordinate-token confidence, which focuses exclusively on these informative tokens:
\begin{equation}
c_{\text{coord}}(y \mid x_{img}, x_{ins}) =
\frac{1}{k} \sum_{i \in \text{coord}} P(y_i \mid y_1, \dots, y_{i-1}, x_{img}, x_{ins})
\label{eq:coord confidence}
\end{equation}
where $k$ represents the number of coordinate tokens. As shown in \cref{fig:intro_ans} right, this metric effectively distinguishes correct from incorrect predictions. Given an image–instruction pair $(x_{img}, x_{ins})$, we generate multiple predictions through parallel sampling and select the one with highest coordinate-token confidence as our pseudo-label:
\begin{equation}
\hat{y} = \arg\max_{y \in \mathcal{Y}} c_{\text{coord}}(y \mid x_{img}, x_{ins}), \quad
\hat{p} = \text{extract}(\hat{y})
\label{eq:p hat}
\end{equation}
where $\hat{y}$ represents the optimal response selected by coordinate-token confidence, and $\mathcal{Y}$ denotes the collection consisting of generated responses. The function $\text{extract}(y)$ is defined to obtain a point from the response $y$. $\hat{p}$ indicates the optimal predicted point, i.e. the pseudo-label. This confidence-based selection provides a principled basis for self-supervised learning without requiring ground truth annotations.

\subsection{Confidence-Anchored Reinforcement Learning}

To enable reinforcement learning without ground truth annotations, we design a distance-based reward mechanism using our pseudo-labels. For the $i$th sampled prediction $(x_i, y_i)$, given an image with height $H$ and width $W$, we normalize its coordinates as $(x_i/W, y_i/H)$, and compute its Euclidean distance to the pseudo-label and assign binary rewards:
\begin{equation}
R_i =
\begin{cases}
1, & \text{if}\ \text{dist}(p_i, \hat{p}) \le \tau \\
0, & \text{otherwise}
\end{cases}
\label{eq:reward function}
\end{equation}
This formulation exploits the sparse nature of GUI coordinate space where predictions close to our high-confidence pseudo-label are likely correct, while distant predictions are almost certainly wrong. The distance threshold $\tau$ defines the boundary between potentially correct and incorrect predictions.
We then apply Group Relative Policy Optimization (GRPO) \cite{grpo}, which estimates advantages through standardization across multiple samples:
\begin{equation}
\label{eq:advantages}
A_i = \frac{R_i-\text{mean}(\{R_j\}_{j=1}^N)}{\text{std}(\{R_j\}_{j=1}^N)}
\end{equation}
where $N$ denotes the sampling number. The policy optimization follows the standard GRPO objective with clipped probability ratios and KL regularization:
\begin{equation}
\label{eq:grpo}
\mathcal{J}_{\text{GRPO}}(\theta)
= \mathbb{E}_{q, {o_i}}
\frac{1}{G} \sum_{i=1}^G
\left[
\min \left(
r_i(\theta)A_i,
\text{clip}(r_i(\theta), 1{-}\epsilon, 1{+}\epsilon)A_i
\right)
- \beta \mathbb{D}_{\text{KL}}[\pi_\theta | \pi_{\text{ref}}]
\right]
\end{equation}
where the probability ratio $r_i(\theta)$ prevents excessive policy updates, $\beta$ controls the strength of regularization, and $G$ denotes the group size. By treating the model's most confident prediction as a learning target and using spatial proximity to define rewards, CAL transforms GUI grounding into a self-supervised reinforcement learning problem requiring no external annotations.

\subsection{Confidence-Anchored Negative Reinforcement Learning}

While CAL provides a path to label-free training, pseudo-labels inevitably contain errors that can mislead learning, particularly when incorrect predictions are mistakenly rewarded as positive examples. However, the sparse nature of GUI coordinate space offers an opportunity: negative samples are overwhelmingly reliable since predictions far from any reasonable target are almost certainly incorrect. CANL exploits this asymmetry by modifying the advantage computation to use only negative samples:
\begin{equation}
\label{eq:negative advantages}
A_i =
\begin{cases}
0, & \text{if } R_i = 1 \\
\frac{-\text{mean}(\{R_j\}_{j=1}^N)}{\text{std}(\{R_j\}_{j=1}^N)}, & \text{if } R_i = 0
\end{cases}
\end{equation}
This approach zeros out advantages for potentially unreliable positive samples while preserving the learning signal from negative samples. The model thus learns exclusively from what to avoid rather than what to produce, which proves particularly effective in high-resolution GUI tasks where the vast coordinate space makes distant predictions almost certainly incorrect.
As illustrated in \cref{fig:method}, CANL maintains the same pseudo-label generation and reward assignment pipeline as CAL but selectively updates the model using only the most reliable learning signals. This strategy effectively leverages label uncertainty as a catalyst for robust negative learning, yielding superior performance on complex datasets where pseudo-label quality is often vulnerable to noise.

\section{Experiments}

\subsection{Experiments Setup}

\noindent\textbf{Datasets and Benchmarks.}  We evaluate on four benchmarks: ScreenSpot \cite{seeclick} and ScreenSpot-V2 \cite{osatlas}, which have low-resolution interfaces and larger targets, and more challenging benchmark ScreenSpot-Pro \cite{screenspot-pro} and UI-Vision \cite{ui-vision}. For UI-Vision, we only use the Element Grounding subset for training and evaluation, without Layout Grounding and Action Prediction subset. Predictions are correct if they fall within ground truth bounding boxes.

\noindent\textbf{Implementation Details}. We use Qwen-2.5-VL \cite{qwen2.5-vl} (3B/7B) as the base model within the VLM-R1 framework \cite{vlm-r1}. Following \cite{ttrl}, we adopt a test-time training setting where the model is optimized and evaluated on each benchmark independently, obviating the need for a separate training set. All trainings are conducted for 1 epoch, with learning rate 1e-6. To enhance diversity during sampling, we set temperature $T=1.0$, $top\_k=50$, $top\_p=1.0$. KL penalty $\beta$ is set to 0.04. Distance threshold $\tau$ is set to 0.05. We apply Flash Attention2 \cite{flashattention} for training. Following \cite{ttrl}, to reduce computational costs while improve the accuracy of pseudo-labels, we apply downsampling strategy during training. Specifically, we sample 16 responses to construct pseudo labels. For CAL, we randomly downsample 8 responses to compute the advantages. For CANL, we compute the advantages using 16 responses and then select the 8 negative responses farthest from the pseudo-label for policy optimization.

\noindent\textbf{Baselines.} We compare against three categories of methods: (1) Label-required models, including SFT paradigms (e.g., SeeClick \cite{seeclick}, ShowUI \cite{showui}, and UI-TARS \cite{ui-tars}) and RL approaches (e.g., UI-R1 \cite{ui-r1}, GUI-R1 \cite{gui-r1}); (2) Label-free model: GUI-RCPO \cite{gui-rcpo}, which leverages region consistency for training guidance; (3) Proprietary models such as GPT-4o \cite{gpt4o} and Claude Computer Use \cite{claude}. Both our method and GUI-RCPO are trained on test sets, enabling a fair performance comparison. By contrast, comparisons with label-required models mainly highlight our data efficiency.

\subsection{Main Results}

\begin{table}[t]
    \centering
    \caption{Performance comparison on ScreenSpot-V1 and V2. "-" indicates missing values due to unavailable results, unreleased checkpoints, and code. For label-free methods, the optimal and the suboptimal results are \textbf{bolded} and \underline{underlined}, respectively.}
    \resizebox{\textwidth}{!}{ %
        \setlength{\tabcolsep}{6pt}
\small
\begin{tabular}{lccccccccc}
\midrule
\multirow{2}{*}{\bf Model} &
\multirow{2}{*}{\bf GUI Labels} &
\multicolumn{2}{c}{\bf v1 Mobile} &
\multicolumn{2}{c}{\bf v1 Desktop} &
\multicolumn{2}{c}{\bf v1 Web} &
\multirow{2}{*}{\bf v1 Avg.} &
\multirow{2}{*}{\bf v2 Avg.} \\
\cmidrule(lr){3-4} \cmidrule(lr){5-6} \cmidrule(lr){7-8}
 & & Text & Icon & Text & Icon & Text & Icon & \\
\midrule
\rowcolor{gray!15}
\multicolumn{10}{l}{\textit{Proprietary Models}} \\
GPT-4o \cite{gpt4o}& - & 30.5 & 23.2 & 20.6 & 19.4 & 11.1 & 7.8 & 18.8 & 20.1\\
Claude Computer Use \cite{claude}& - & - & - & - & - & - & - & 83.0 & -\\
\rowcolor{gray!15}
\multicolumn{10}{l}{\textit{General Models}} \\
Qwen-2.5-VL-3B \cite{qwen2.5-vl}& 0 & 93.8 & 68.1 & 91.2 & 55.0 & 81.7 & 64.6 & 77.6 & 82.1 \\
Qwen-2.5-VL-7B \cite{qwen2.5-vl}& 0 & 91.9 & 80.8 & 88.1 & 75.7 & 90.0 & 77.7 & 84.9 & 88.1 \\
\rowcolor{gray!15}
\multicolumn{10}{l}{\textit{GUI-specific Models (Label-required)}} \\
CogAgent-18B \cite{cogagent}& 222M & 67.0 & 24.0 & 74.2 & 20.0 & 70.4 & 28.6 & 47.4 & - \\
SeeClick-9.6B \cite{seeclick}& 1M & 78.0 & 52.0 & 72.2 & 30.0 & 55.7 & 32.5 & 53.4 & 55.1 \\
UGround-7B \cite{uground}& 10M & 82.8 & 60.3 & 82.5 & 63.6 & 80.4 & 70.4 & 73.3 & 76.3 \\
OS-Atlas-7B \cite{osatlas}& 13M & 93.0 & 72.9 & 91.8 & 62.9 & 90.9 & 74.3 & 82.5 & - \\
ShowUI-2B \cite{showui}& 256K & 92.3 & 75.5 & 76.3 & 61.1 & 81.7 & 63.6 & 75.1 & 77.3 \\
Aguvis-72B \cite{aguvis}& 1M & 94.5 & 85.2 & 95.4 & 77.9 & 91.3 & 85.9 & 89.2 & - \\
UI-TARS-7B \cite{ui-tars}& 18.4M & 94.5 & 85.2 & 95.9 & 85.7 & 90.0 & 83.5 & 89.5 & 91.6 \\
UI-TARS-72B \cite{ui-tars}& 18.4M & 94.9 & 82.5 & 89.7 & 88.6 & 88.7 & 85.0 & 88.4 & 90.3 \\
GUI-Actor-7B \cite{gui-actor}& 9.6M & 94.9 & 82.1 & 91.8 & 80.0 & 91.3 & 85.4 & 88.3 & 92.1 \\
Jedi-7B \cite{osworldg_jedi}& 4M & - & - & - & - & - & - & - & 91.7 \\
UI-R1-3B \cite{ui-r1}& 136 & 95.6 & 84.7 & 90.2 & 59.3 & 85.2 & 73.3 & 83.3 & 85.4 \\
GUI-R1-7B \cite{gui-r1}& 3K & - & - & 91.8 & 73.6 & 91.3 & 75.7 & - & - \\
InfiGUI-R1-3B \cite{infigui-r1}& 32K & 97.1 & 81.2 & 94.3 & 77.1 & 91.7 & 77.6 & 87.5 & - \\
SE-GUI-7B \cite{gui-se}& 3K & - & - & - & - & - & - & 88.2 & 90.3 \\
GuirlVG-7B \cite{guirlvg}& 5.2K & 96.0 & 84.7 & 92.8 & 80.0 & 92.6 & 85.9 & 88.7 & 91.9 \\
\midrule
\rowcolor{gray!15}
\multicolumn{10}{l}{\textit{GUI-specific Models (Label-free)}} \\
GUI-RCPO-7B \cite{gui-rcpo}& 0 & - & - & - & - & - & - & 86.6 & \underline{88.9} \\
\rowcolor{gray!15}
\multicolumn{10}{l}{\textit{Ours}} \\
CAL-3B & 0 & \underline{96.7} & 78.6 & \underline{95.4} & 67.9 & 87.8 & 72.8 & 84.6 & \underline{88.9}\\
CANL-3B & 0 & 96.0 & \underline{79.0} & \textbf{96.4} & 66.4 & 87.0 & 73.8 & 84.5 & 88.5\\
CAL-7B & 0 & \textbf{97.1} & \textbf{87.3} & 86.1 & \underline{80.7} & \textbf{91.7} & \underline{83.0} & \underline{88.6} & \textbf{92.1} \\
CANL-7B & 0 & \underline{96.7} & \textbf{87.3} & 88.6 & \textbf{82.1} & \underline{91.3} & \textbf{84.0} & \textbf{89.2} & \textbf{92.1} \\
\bottomrule
\end{tabular}

    }
    \label{tab:v1_v2}
\end{table}

\begin{table}[htb]
    \centering
    \caption{Performance comparison on ScreenSpot-Pro. For label-free methods, the optimal and the suboptimal results are \textbf{bolded} and \underline{underlined}, respectively.}
    \resizebox{\textwidth}{!}{ %
        \setlength{\tabcolsep}{2pt}
\small
\begin{tabular}{lcccccccccccccc}
\toprule
\multirow{2}{*}{\textbf{Model}} &
\multirow{2}{*}{\bf GUI Labels} &
\multicolumn{2}{c}{\textbf{CAD}} &
\multicolumn{2}{c}{\textbf{Dev}} &
\multicolumn{2}{c}{\textbf{Creative}} &
\multicolumn{2}{c}{\textbf{Scientific}} &
\multicolumn{2}{c}{\textbf{Office}} &
\multicolumn{2}{c}{\textbf{OS}} &
\multirow{2}{*}{\textbf{Avg.}}\\
\cmidrule(lr){3-4} \cmidrule(lr){5-6} \cmidrule(lr){7-8} \cmidrule(lr){9-10}
\cmidrule(lr){11-12} \cmidrule(lr){13-14}
 & & Text & Icon & Text & Icon & Text & Icon & Text & Icon & Text & Icon & Text & Icon \\
\midrule
\rowcolor{gray!15}
\multicolumn{15}{l}{\textit{Proprietary Models}} \\
GPT-4o \cite{gpt4o}& - & 2.0 & 0.0 & 1.3 & 0.0 & 1.0 & 0.0 & 2.1 & 0.0 & 1.1 & 0.0 & 0.0 & 0.0 & 0.8 \\
Claude Computer Use \cite{claude}& - & 14.5 & 3.7 & 22.0 & 3.9 & 25.9 & 3.4 & 33.9 & 15.8 & 30.1 & 16.3 & 11.0 & 4.5 & 17.1 \\
\rowcolor{gray!15}
\multicolumn{15}{l}{\textit{General Models}} \\
Qwen-2.5-VL-3B \cite{qwen2.5-vl}& 0 & 9.1 & 7.3 & 22.1 & 1.4 & 26.8 & 2.1 & 38.2 & 7.3 & 33.9 & 15.1 & 10.3 & 1.1 & 16.1\\
Qwen-2.5-VL-7B \cite{qwen2.5-vl}& 0 & 13.7 & 7.8 & 44.2 & 6.9 & 28.8 & 10.5 & 48.6 & 5.5 & 46.9 & 15.1 & 31.8 & 12.4 & 24.9\\
\rowcolor{gray!15}
\multicolumn{15}{l}{\textit{GUI-specific Models (Label-required)}} \\
SeeClick-9.6B \cite{seeclick}& 1M & 2.5 & 0.0 & 0.6 & 0.0 & 1.0 & 0.0 & 3.5 & 0.0 & 1.1 & 0.0 & 2.8 & 0.0 & 1.1\\
CogAgent-18B \cite{cogagent}& 222M & 7.1 & 3.1 & 14.9 & 0.7 & 9.6 & 0.0 & 22.2 & 1.8 & 13.0 & 0.0 & 5.6 & 0.0 & 7.7\\
Aria-UI \cite{aria-ui}& 17.6M & 7.6 & 1.6 & 16.2 & 0.0 & 23.7 & 2.1 & 27.1 & 6.4 & 20.3 & 1.9 & 4.7 & 0.0 & 11.3\\
OS-Atlas-7B \cite{osatlas}& 13M & 12.2 & 4.7 & 33.1 & 1.4 & 28.8 & 2.8 & 37.5 & 7.3 & 33.9 & 5.7 & 27.1 & 4.5 & 18.9\\
ShowUI-2B \cite{showui}& 256K & 2.5 & 0.0 & 16.9 & 1.4 & 9.1 & 0.0 & 13.2 & 7.3 & 15.3 & 7.5 & 10.3 & 2.2 & 7.7\\
UGround-7B \cite{uground}& 10M & 14.2 & 1.6 & 26.6 & 2.1 & 27.3 & 2.8 & 31.9 & 2.7 & 31.6 & 11.3 & 17.8 & 0.0  & 16.5\\
ZonUI-3B \cite{zonui}& 24K & 31.9 & 15.6 & 24.6 & 6.2 & 40.9 & 7.6 & 54.8 & 18.1 & 57.0 & 26.4 & 19.6 & 7.8 & 28.7\\
UI-R1-3B \cite{ui-r1}& 136 & 11.2 & 6.3 & 22.7 & 4.1 & 27.3 & 3.5 & 42.4 & 11.8 & 32.2 & 11.3 & 13.1 & 4.5 & 17.8\\
GUI-R1-7B \cite{gui-r1}& 3K & 23.9 & 6.3 & 49.4 & 4.8 & 38.9 & 8.4 & 55.6 & 11.8 & 58.7 & 26.4 & 42.1 & 16.9 & 31.0\\
UI-Ins-32B \cite{ui-ins}& 316K & 51.8 & 29.7 & 83.1 & 26.9 & 69.7 & 18.9 & 83.3 & 34.5 & 88.7 & 50.9 & 70.1 & 34.8 & 57.0 \\
\midrule
\rowcolor{gray!15}
\multicolumn{15}{l}{\textit{GUI-specific Models (Label-free)}} \\
GUI-RCPO-7B \cite{gui-rcpo}& 0 &- &- &- &- &- &- &- &- &- &- &- &- &25.9\\
\rowcolor{gray!15}
\multicolumn{15}{l}{\textit{Ours}} \\
CAL-3B & 0 & \underline{35.0} & 9.4 & 44.2 & 3.4 & \textbf{46.5} & 4.9 & \underline{60.4} & \underline{18.1} & 53.1 & 20.8 & \underline{38.3} & 7.9 & 32.1 \\
CANL-3B & 0 & \textbf{39.1} & 7.8 & 42.9 & 4.8 & \underline{46.0} & 3.5 & 58.3 & \textbf{20.9} & 55.9 & \underline{22.6} & \underline{38.3} & 7.9 & 32.7 \\
CAL-7B & 0 & 25.4 & \textbf{11.0} & \underline{51.3} & \underline{6.9} & 38.9 & \textbf{9.8} & \underline{60.4} & 15.5 & \textbf{64.7} & 19.2 & 37.4 & \textbf{19.2} & \underline{32.7} \\
CANL-7B & 0 & 29.4 & \underline{10.9} & \textbf{53.2} & \textbf{9.0} & 37.4 & \underline{8.4} & \textbf{63.2} & 14.5 & \underline{62.1} & \textbf{26.4} & \textbf{40.2} & \underline{15.7} & \textbf{33.8} \\
\bottomrule
\end{tabular}

    }
    \label{tab:pro}
\end{table}

\begin{figure}[!t]
    \centering
    \begin{subfigure}[b]{0.48\textwidth}
        \centering
        \includegraphics[width=\textwidth]{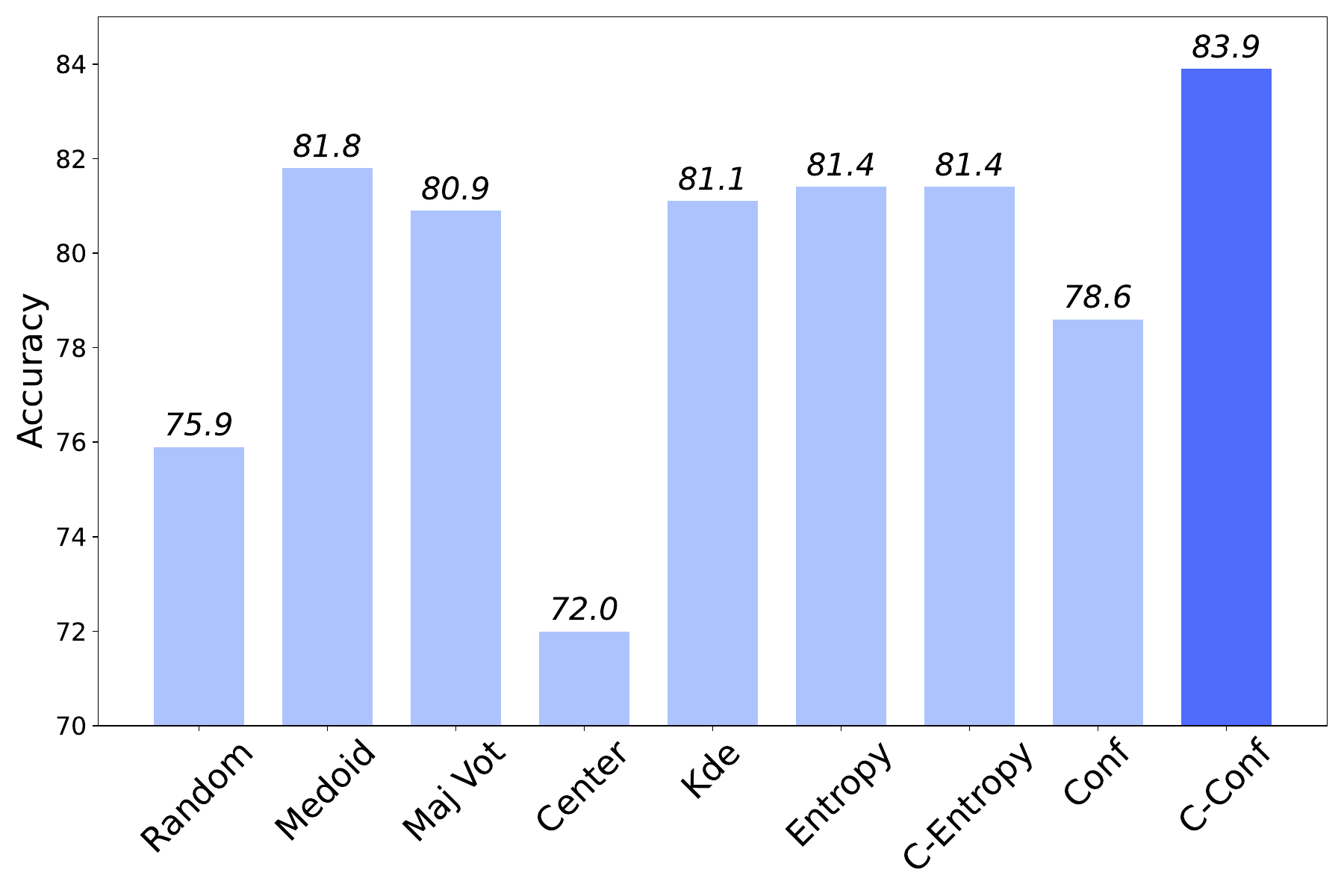}
        \label{fig:method_ablation}
    \end{subfigure}
    \hfill
    \begin{subfigure}[b]{0.48\textwidth}
        \centering
        \includegraphics[width=\textwidth]{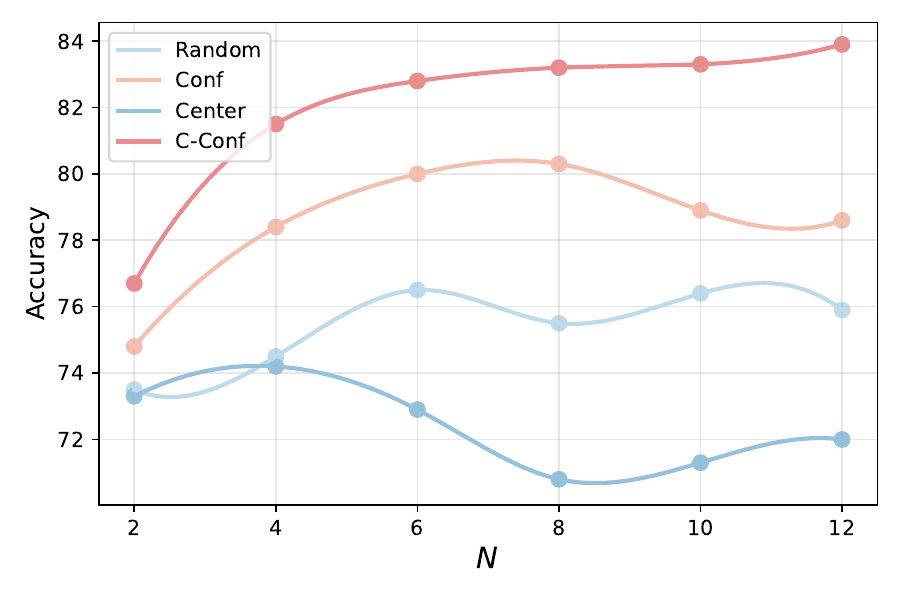}
        \label{fig:rollout_num}
    \end{subfigure}
    \caption{Performance of different pseudo-label construction methods. \textbf{Left}: Comparison of pseudo-label construction methods with sampling number $N$=12. \textbf{Right}: Pseudo-label quality scales with sampling budget across four strategies.}
    \label{fig: pseudo method}
\end{figure}

\begin{table}[t]
    \centering
    \caption{Performance comparison on UI-Vision. For label-free methods, the optimal and the suboptimal results are \textbf{bolded} and \underline{underlined}, respectively.}
    \resizebox{\textwidth}{!}{ %
        \setlength{\tabcolsep}{2pt}
\small
\begin{tabular}{lccccccccccc}
\toprule
\multirow{2}{*}{\bf Model} & 
\multirow{2}{*}{\bf GUI Labels} &
\multicolumn{6}{c}{\bf Grouped by Category} & 
\multicolumn{3}{c}{\bf Grouped by Setting} & 
\multirow{2}{*}{\bf Overall} \\
\cmidrule(r){3-8} \cmidrule(r){9-11}
 & & Edu. & Browser & Dev. & Prod. & Creative & Entert. & Basic & Func. & Spatial \\
\midrule
\rowcolor{gray!15}
\multicolumn{12}{l}{\textit{Proprietary Models}} \\
GPT-4o \cite{gpt4o}& - & 1.5 & 0.0 & 2.2 & 1.1 & 0.8 & 4.2 & 1.6 & 1.5 & 1.0 & 1.4 \\
Claude Computer Use \cite{claude}& - & 6.1 & 9.8 & 8.0 & 9.4 & 7.7 & 8.3 & 9.5 & 7.7 & 7.6 & 8.3 \\
\rowcolor{gray!15}
\multicolumn{12}{l}{\textit{General Models}} \\
Qwen-2.5-VL-3B \cite{qwen2.5-vl}& 0 & 7.6 & 22.4 & 15.4 & 12.8 & 6.6 & 33.9 & 18.6 & 13.8 & 4.3 & 12.0\\
Qwen-2.5-VL-7B \cite{qwen2.5-vl}& 0 & 11.1 & 37.1 & 18.1 & 15.4 & 9.6 & 29.7 & 20.0 & 18.6 & 7.1 & 15.0\\
\rowcolor{gray!15}
\multicolumn{12}{l}{\textit{GUI-specific Models (Label-required)}} \\
SeeClick-9.6B \cite{seeclick}& 1M & 4.2 & 13.3 & 7.3 & 4.3 & 4.0 & 11.0 & 9.4 & 4.7 & 2.1 & 5.4 \\
ShowUI-2B \cite{showui}& 256K & 3.7 & 13.3 & 7.5 & 6.5 & 2.5 & 15.6 & 8.1 & 7.7 & 2.1 & 5.9 \\
CogAgent-9B \cite{cogagent}& 222M & 8.7 & 11.2 & 8.6 & 10.3 & 5.6 & 15.6 & 12.0 & 12.2 & 2.6 & 8.9 \\
OSAtlas-7B \cite{osatlas}& 13M & 8.7 & 16.8 & 10.3 & 9.2 & 5.6 & 16.2 & 12.2 & 11.2 & 3.7 & 9.0 \\
AriaUI \cite{aria-ui}& 17.6M & 9.0 & 18.9 & 11.2 & 10.4 & 6.5 & 19.3 & 12.2 & 14.0 & 4.0 & 10.1 \\
UGround-v1-7B \cite{uground}& - & 10.4 & 28.7 & 17.5 & 12.2 & 8.6 & 18.2 & 15.4 & 17.1 & 6.3 & 12.9 \\
Aguvis-7B \cite{aguvis}& 1M & 13.1 & 30.8 & 17.1 & 12.1 & 9.6 & 24.0 & 17.8 & 18.3 & 5.1 & 13.7 \\
UI-TARS-7B \cite{ui-tars}& 18.4M & 14.2 & 35.0 & 19.7 & 18.3 & 11.1 & 38.5 & 20.1 & 24.3 & 8.4 & 17.6 \\
\midrule
\rowcolor{gray!15}
\multicolumn{12}{l}{\textit{Ours}} \\

CAL-3B & 0 & 15.1 & 32.9 & 20.6 & 18.1 & 10.4 & 37.0 & 24.9 & 22.1 & 5.7 & 17.2\\
CANL-3B & 0 & 16.0 & 33.6 & 21.6 & \underline{19.4} & \underline{12.0} & 39.1 & \underline{26.4} & \underline{23.8} & 6.5 & 18.5\\
CAL-7B & 0 & \textbf{17.6} & \textbf{42.7} & \underline{22.0} & 18.9 & 10.9 & \underline{41.1} & 25.6 & 22.9 & \underline{8.4} & \underline{18.6}\\
CANL-7B & 0 & \underline{17.3} & \underline{38.5} & \textbf{24.1} & \textbf{20.9} & \textbf{12.4} & \textbf{44.8} & \textbf{27.5} & \textbf{25.1} & \textbf{8.8} & \textbf{20.1}\\
\bottomrule
\end{tabular}

    }
    \label{tab:ui-vision}
\end{table}

\noindent\textbf{Our label-free methods achieve competitive performance without any annotations.} \Cref{tab:v1_v2} demonstrates substantial improvements over vanilla base models on ScreenSpot and ScreenSpot-V2, with absolute gains ranging from 4.0\% to 7\%. On ScreenSpot-V2, CANL-7B achieves 92.1\%, improving 4.0\% over the base Qwen-2.5-VL-7B, outperforming GUI-RCPO-7B (88.9\%), proving the reward signal of our methods is more reliable. Meanwhile, CAL-7B matches GUI-Actor-7B (92.1\%) which uses 9.6M labels. At the 3B scale, CANL improves the base model from 77.6\% to 84.5\% (+6.9\%), approaching InfiGUI-R1-3B (87.5\%) which requires 32K labels, validating that coordinate-token confidence provides sufficient supervision.

\noindent\textbf{Negative learning dominates on challenging high-resolution benchmarks.} As shown in \cref{tab:pro} and \cref{tab:ui-vision}, the performance gap between CAL and CANL reveals a critical pattern: as difficulty increases, exclusive negative learning becomes increasingly advantageous. On ScreenSpot-Pro, CANL-7B achieves 33.8\%, improving 1.1\% over CAL-7B, 8.9\% over the base model, and GUI-RCPO-7B (25.9\%) by a notable margin. This advantage extends to UI-Vision where CANL-7B reaches 20.1\%, surpassing CAL by 1.5\%. The systematic superiority of CANL confirms that on high-resolution and challenging tasks, negative samples provide more reliable learning signals than potentially incorrect positive samples derived from pseudo-labels.

\begin{figure}[!t]
    \centering
    \begin{subfigure}[b]{0.48\textwidth}
        \centering
        \includegraphics[width=\textwidth]{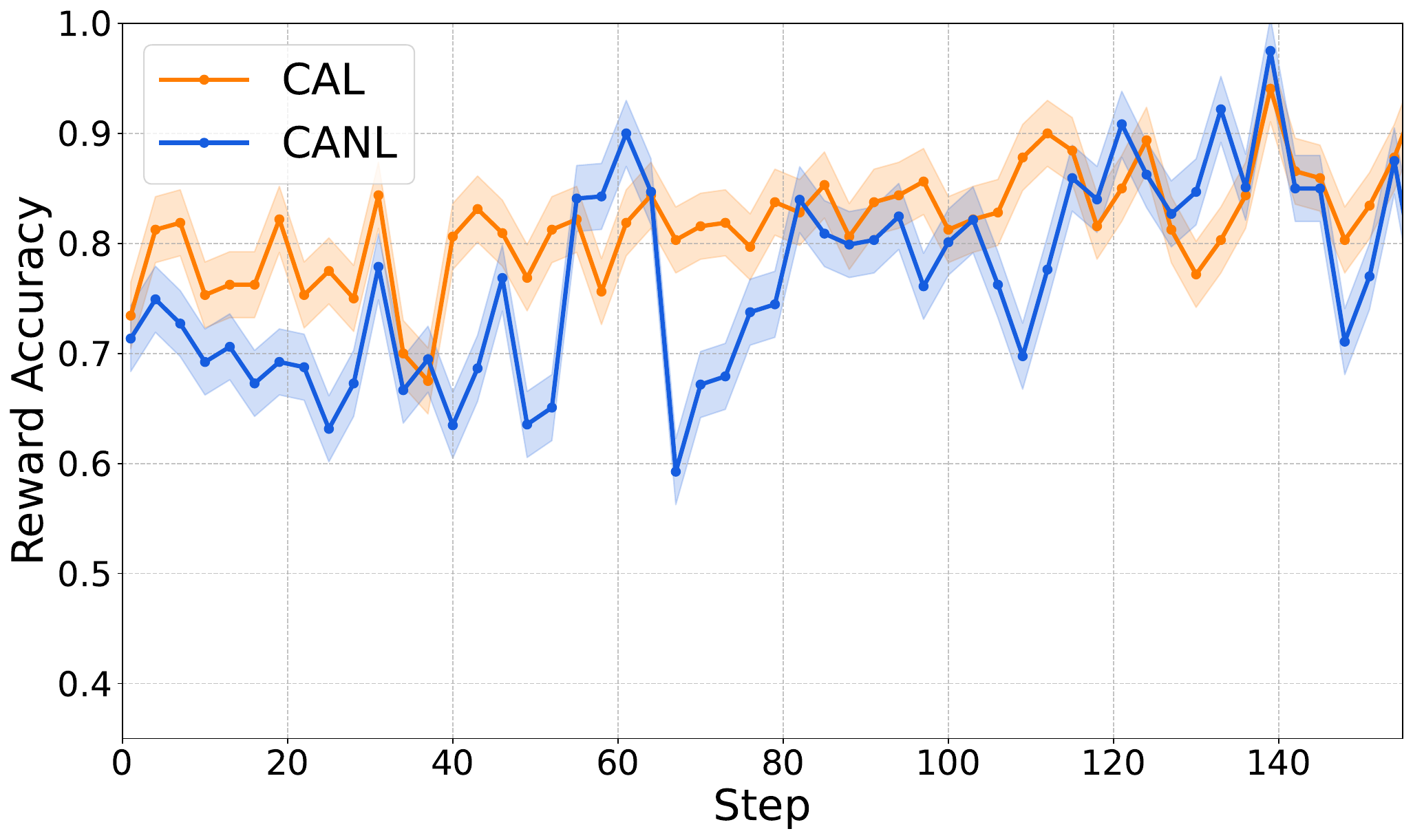}  %
        \label{fig:v2 reward acc}
    \end{subfigure}
    \hfill  %
    \begin{subfigure}[b]{0.48\textwidth}
        \centering
        \includegraphics[width=\textwidth]{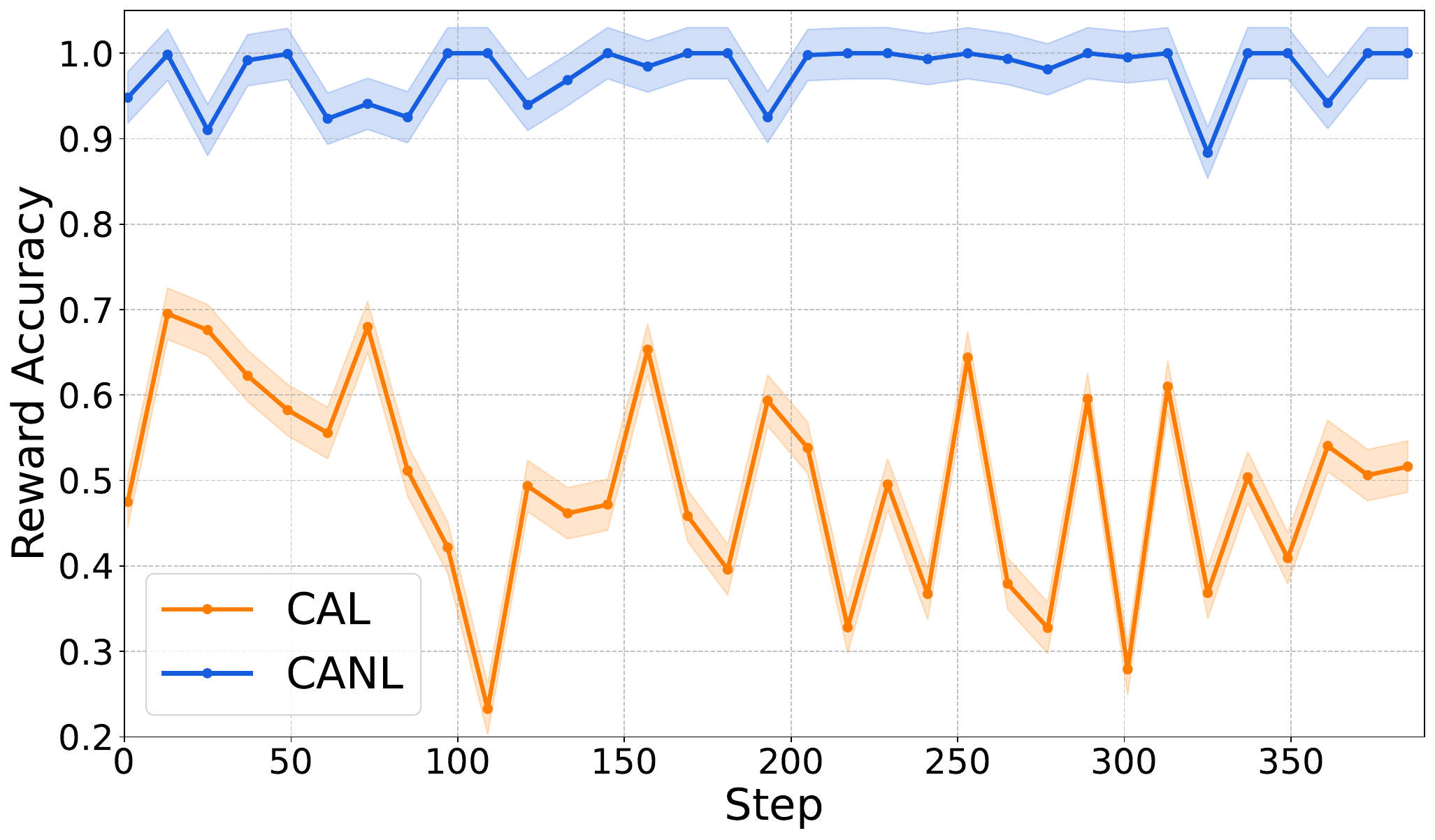}  %
        \label{fig:pro reward acc}
    \end{subfigure}
    \caption{\textbf{Left}: Reward accuracy on ScreenSpot-V2. \textbf{Right}: Reward accuracy on ScreenSpot-Pro.}  %
    \label{fig:reward acc}
\end{figure}

\subsection{In-depth Analysis}
\noindent\textbf{Coordinate-token confidence outperforms alternative pseudo-labeling strategies.} \Cref{fig: pseudo method} evaluates different test-time scaling methods to generate pseudo-label on ScreenSpot-V2. Detailed descriptions of these methods are provided in Appendix \ref{sec:pseudo_baseline_intro}. Our proposed coordinate-token confidence (C-Conf) achieves 83.9\% accuracy with $N=12$ samples, substantially exceeding alternatives. The 5.3\% gain over standard confidence validates our core insight that coordinate tokens carry the primary uncertainty signal in GUI grounding, as formatting and descriptive tokens exhibit consistently high probabilities regardless of prediction quality. Moreover, this superiority persists across varying sampling budgets: C-Conf maintains its advantage from minimal sampling ($N=2$: 76.7\% vs 74.8\% for standard confidence) to larger ensembles ($N=12$: 83.9\% vs 78.6\%). Notably, while spatial methods like Center collapse at higher sampling rates due to averaging effects, C-Conf shows monotonic improvement, confirming that confidence-based selection scales effectively with computational budget.

\noindent\textbf{Reward reliability validates the asymmetric quality of positive and negative samples.} \Cref{fig:reward acc} quantifies the actual correctness of our binary reward assignments by comparing against ground truth labels. On ScreenSpot-V2 with larger bounding boxes, CANL's reward accuracy falls slightly below CAL's because some correct predictions lying outside threshold $\tau$ are misclassified as negative. However, CANL maintains near-perfect negative sample reliability ($>$0.95) while CAL's mixed positive-negative accuracy hovers around 0.5. This empirical validation confirms our core hypothesis that high-resolution GUI tasks naturally provide accurate negative signals even without labels, as the vast coordinate space makes random distant predictions almost certainly incorrect.

\begin{wraptable}{r}{0.35\textwidth}
\centering
\small
\setlength{\tabcolsep}{4pt} 

\caption{Distance threshold $\tau$ sensitivity analysis.}

\begin{tabular}{ccc}
    \toprule
    $\tau$ & SS-V2 & SS-Pro \\ 
    \midrule
    \rowcolor{gray!15}
    \multicolumn{3}{l}{\textit{CAL}} \\
    0.01 & 87.7 & 33.3 \\
    0.03 & 88.4 & 33.2 \\
    0.05 & \textbf{88.9} & 33.2 \\
    0.07 & \textbf{88.9} & 30.1 \\
    0.1  & 86.9 & 30.7 \\
    \midrule
    \rowcolor{gray!15}
    \multicolumn{3}{l}{\textit{CANL}} \\
    0.01 & 87.5 & 30.2 \\
    0.03 & 87.8 & 30.8 \\
    0.05 & \underline
    {88.5} & \textbf{33.7} \\
    0.07 & 88.4 & 32.7 \\
    0.1  & \underline{88.5} & \underline{33.5} \\
    \bottomrule
\end{tabular}
\label{tab:distance_thres}
\end{wraptable}

\noindent\textbf{Distance threshold sensitivity reveals fundamental differences between positive and negative learning.} \Cref{tab:distance_thres} examines how the threshold $\tau$ affects both methods across datasets. On ScreenSpot-V2, CAL exhibits strong sensitivity with performance degrading from 88.9\% at $\tau=0.05$ to 86.9\% at $\tau=0.1$, reflecting its reliance on accurate positive sample identification. Conversely, CANL maintains remarkable stability at 88.5\%±1.0\% across all thresholds. This pattern intensifies on ScreenSpot-Pro where CAL's performance drops precipitously from 33.2\% to 30.1\% as $\tau$ increases, while CANL remain stable. This invariance stems from a key property of sparse coordinate spaces: predictions far from any reasonable target remain unambiguously incorrect as distance threshold increases. While positive samples require careful boundary definition to avoid including incorrect predictions, negative samples beyond any plausible threshold provide consistently reliable learning signals, making CANL robust to hyperparameter.

\begin{figure}[t]
    \centering
    \begin{subfigure}[b]{0.48\textwidth}
        \centering
        \includegraphics[width=\textwidth]{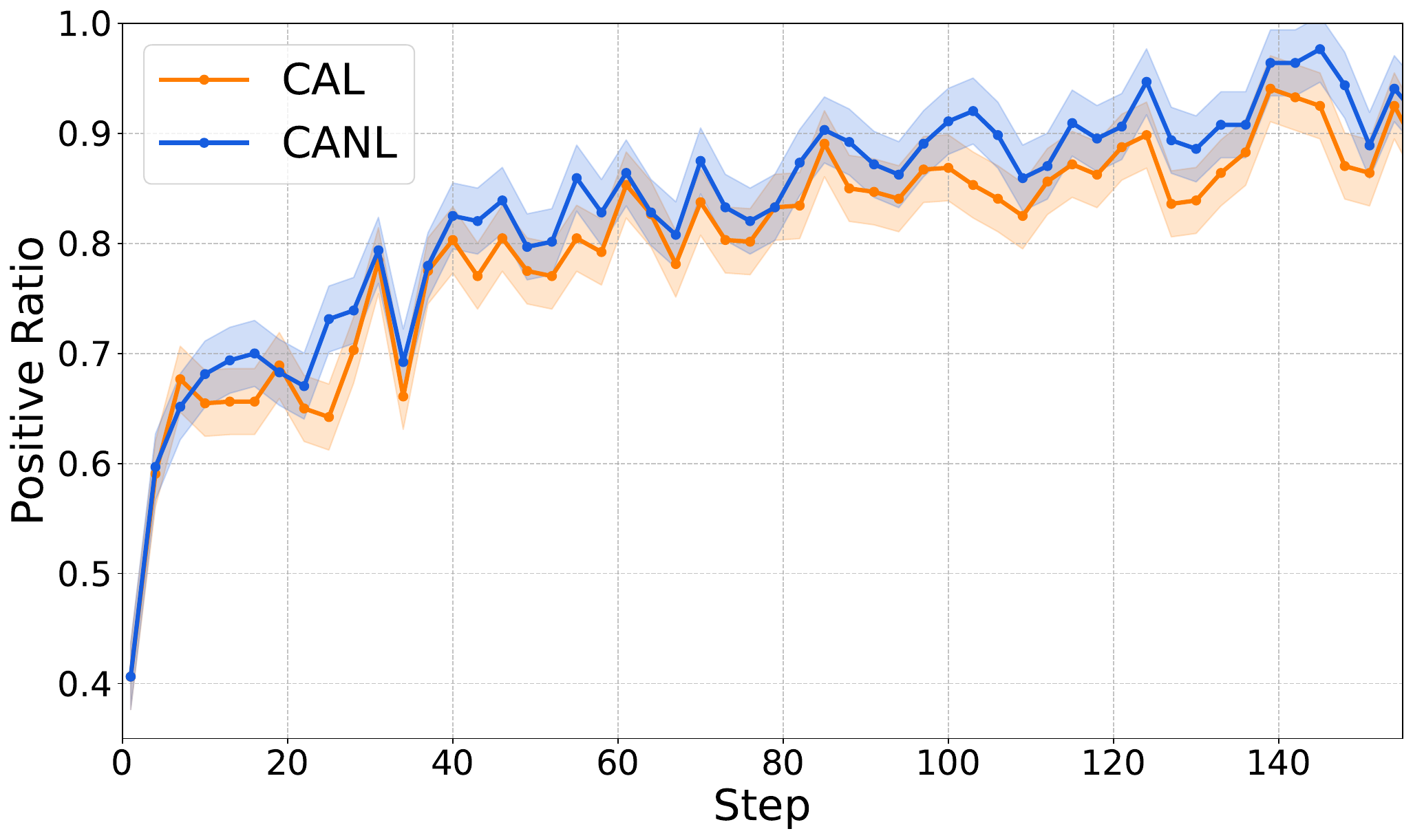}
        \label{fig:v2 ps ratio}
    \end{subfigure}
    \hfill
    \begin{subfigure}[b]{0.48\textwidth}
        \centering
        \includegraphics[width=\textwidth]{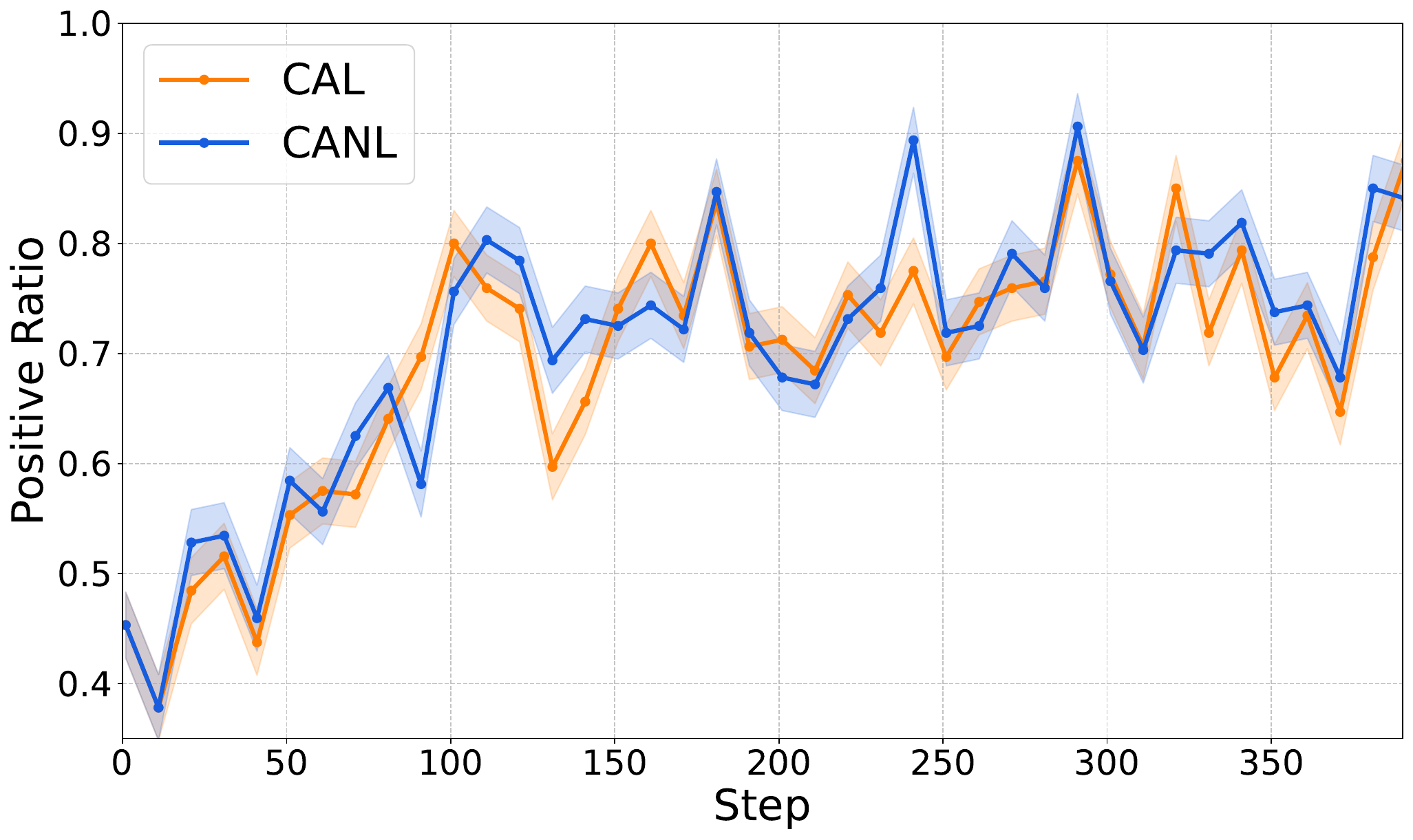}
        \label{fig:pro ps ratio}
    \end{subfigure}
    \caption{\textbf{Left}: Positive Ratio on ScreenSpot-V2. \textbf{Right}: Positive Ratio on ScreenSpot-Pro.}  %
    \label{fig:ps ratio}
\end{figure}

\noindent\textbf{Training dynamics reveal distinct convergence patterns across difficulty levels.} \Cref{fig:ps ratio} tracks the evolution of positive ratio, defined as the fraction of samples within distance threshold $\tau$ from the pseudo-label. On ScreenSpot-V2, both methods exhibit monotonic increases, indicating progressive prediction concentration around confident regions. ScreenSpot-Pro presents markedly different dynamics: slower convergence with higher variance, plateauing around 0.8 rather than continuing upward. This divergence reflects the inherently greater challenge of high-resolution grounding where precise localization becomes critical. Notably, CANL consistently maintains higher positive ratios than CAL across both settings, suggesting that learning exclusively from negative samples paradoxically helps models develop more concentrated predictions, possibly by more effectively pruning incorrect hypotheses from the prediction space.

\begin{figure}[t]
    \centering
    \begin{subfigure}[b]{0.48\textwidth}
        \centering
        \includegraphics[width=\textwidth]{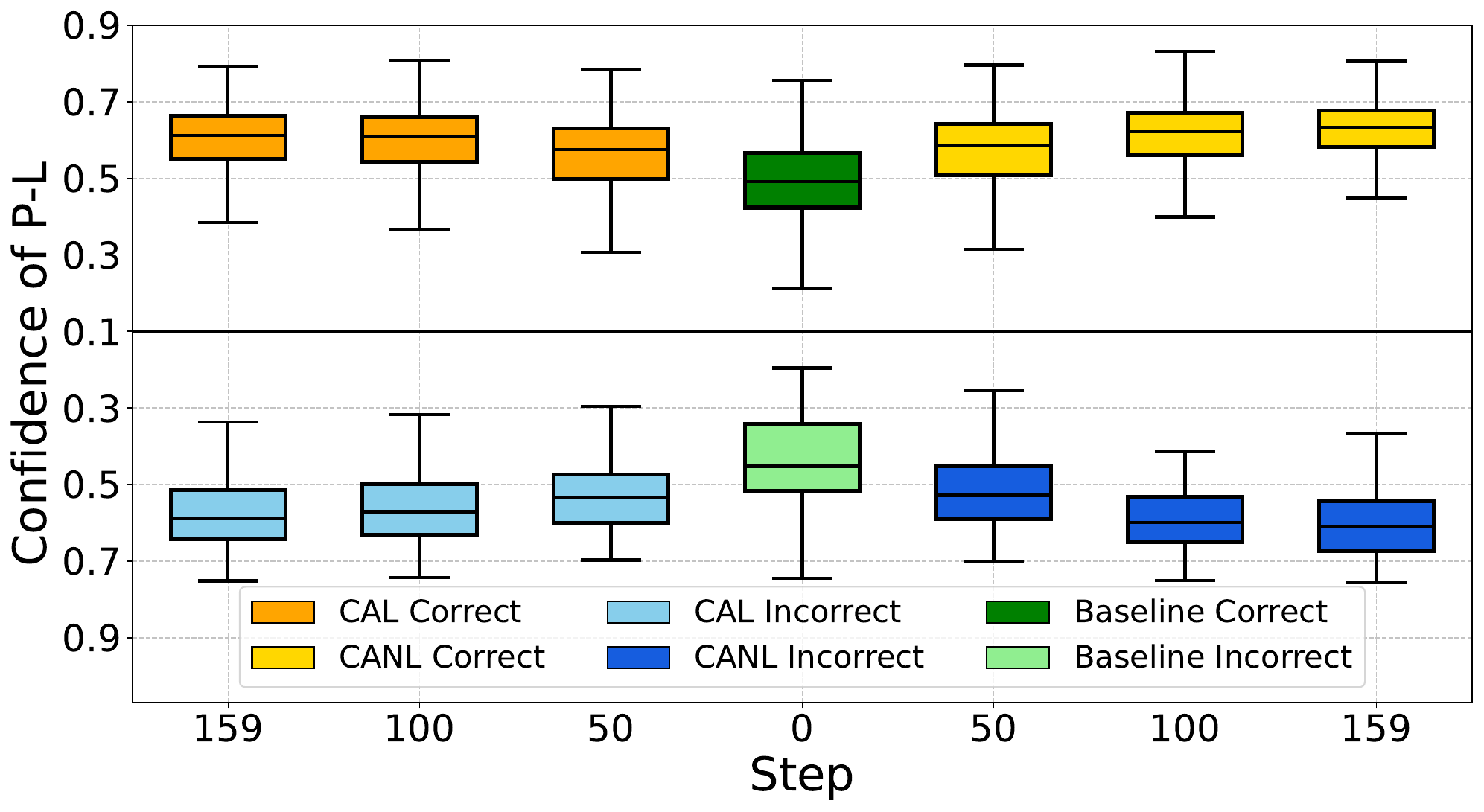}  %
        \label{fig:confidence_box}
    \end{subfigure}
    \hfill  %
    \begin{subfigure}[b]{0.48\textwidth}
        \centering
        \includegraphics[width=\textwidth]{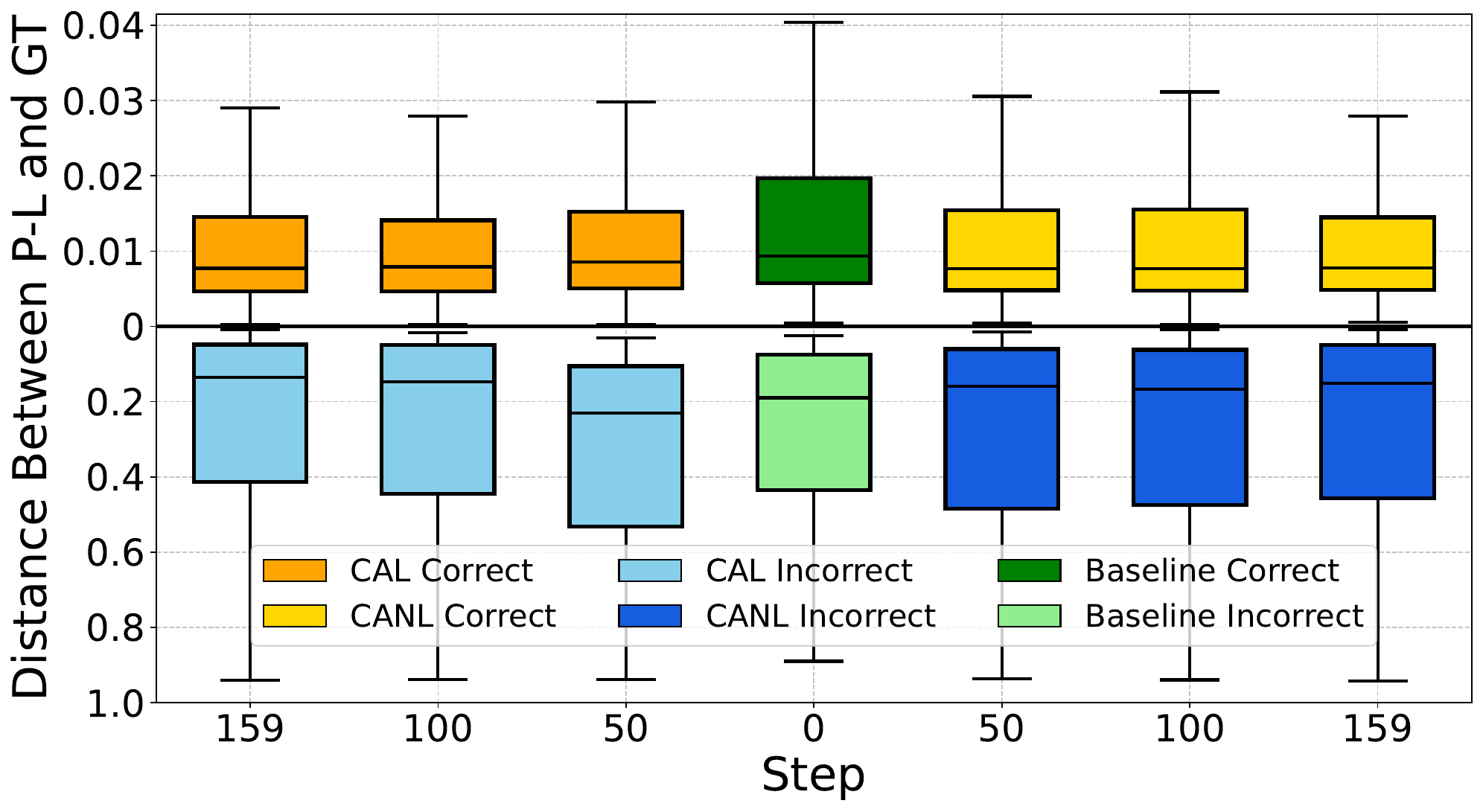}  %
        \label{fig:distance_box}
    \end{subfigure}
    \caption{\textbf{Left}: Confidence of pseudo-label on ScreenSpot-V2. \textbf{Right}: Distance between the pseudo-label and the center of ground truth bounding box on ScreenSpot-V2.}  %
    \label{fig:confidence_distance_box}
\end{figure}

\noindent\textbf{Pseudo-label characteristics reveal both successes and inherent limitations.} \Cref{fig:confidence_distance_box} provides deeper insights into pseudo-label behavior. The left panel shows coordinate-token confidence increases for both correct and incorrect pseudo-labels during training, with correct predictions maintaining a consistent advantage. This convergence suggests models become increasingly confident regardless of accuracy, potentially limiting further improvements without external supervision. The right panel reveals a striking bimodal distribution: correct pseudo-labels cluster within 0.02 normalized distance of ground truth centers, while incorrect ones remain at ~0.40 distance throughout training. This binary pattern validates our distance-based reward design and explains why negative samples are overwhelmingly reliable in sparse coordinate spaces. The persistence of distant incorrect pseudo-labels indicates certain challenging cases remain beyond the model's capability without ground truth, representing an inherent limitation of label-free learning.

\begin{figure}[t]
    \centering
    \includegraphics[width=0.97\linewidth]{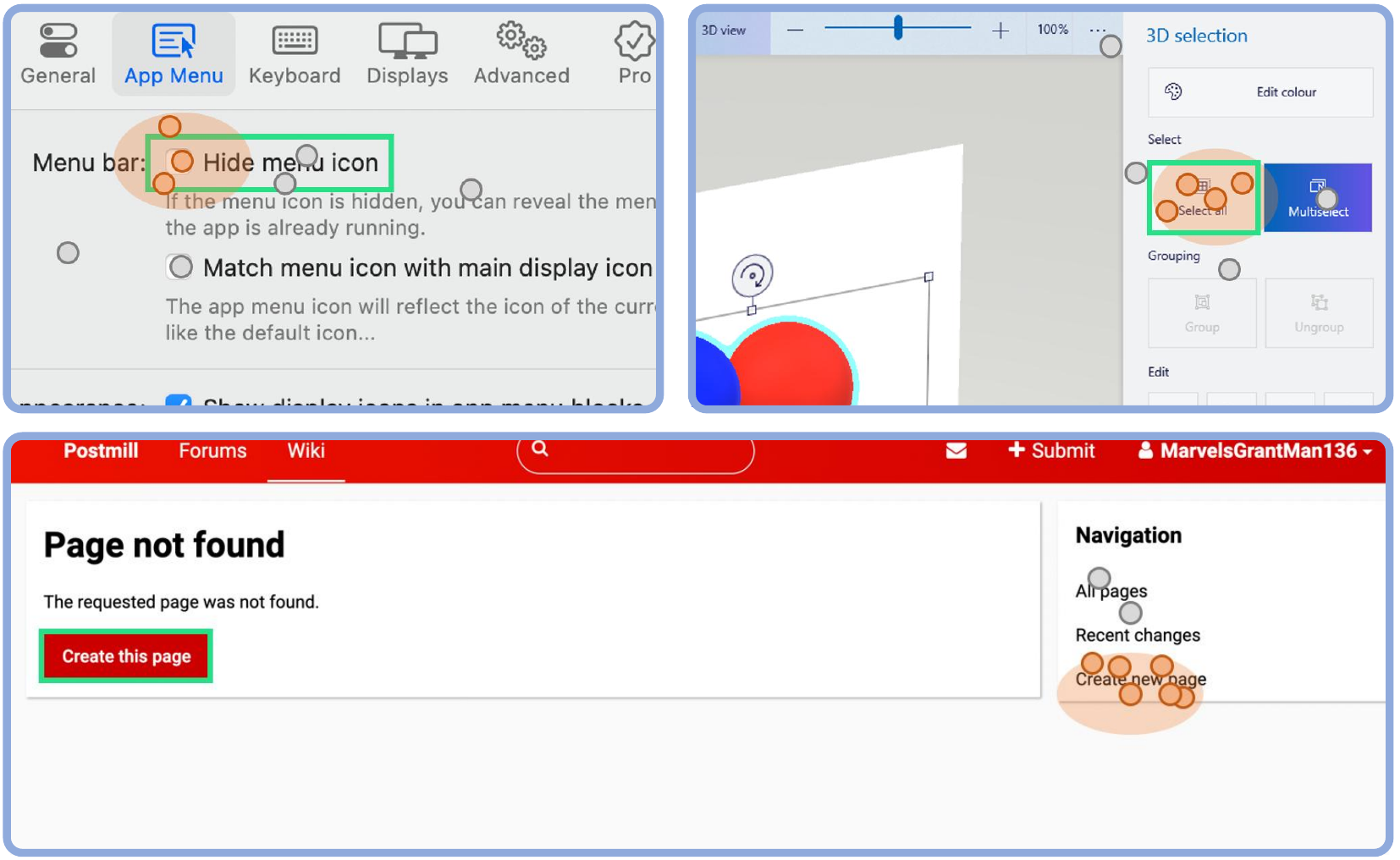}
    \caption{Cropped images of different cases. \textbf{Top-Left}: Geometric mismatch. \textbf{Top-Right}: Good alignment. \textbf{Bottom}: Misplaced pseudo-label.}
    \label{fig:reward estimation case}
\end{figure}

\noindent\textbf{Reward estimation cases demonstrate the advantages of negative learning.} To assess the reliability of our distance-based reward assignment, we analyze three common scenarios during pseudo-label generation, shown in ~\cref{fig:reward estimation case}: \textbf{Geometric mismatch}: The pseudo-label is within the target bounding box, but the ground truth’s elongated shape does not match the region generated by the pseudo-labels. \textbf{Good alignment}: The pseudo-label and reward region align well with the target, resulting in accurate reward estimation. \textbf{Misplaced pseudo-label}: The pseudo-label is far from the target, causing all samples in the reward region to be incorrectly rewarded, though negative samples remain correctly classified. These cases highlight the varying reliability of positive samples under different pseudo-label conditions and explain why CANL’s focus on negative samples yields more robust learning signals.

\begin{figure}[!h]
    \centering
    \begin{subfigure}[b]{0.32\textwidth}
        \centering
        \includegraphics[width=\textwidth]{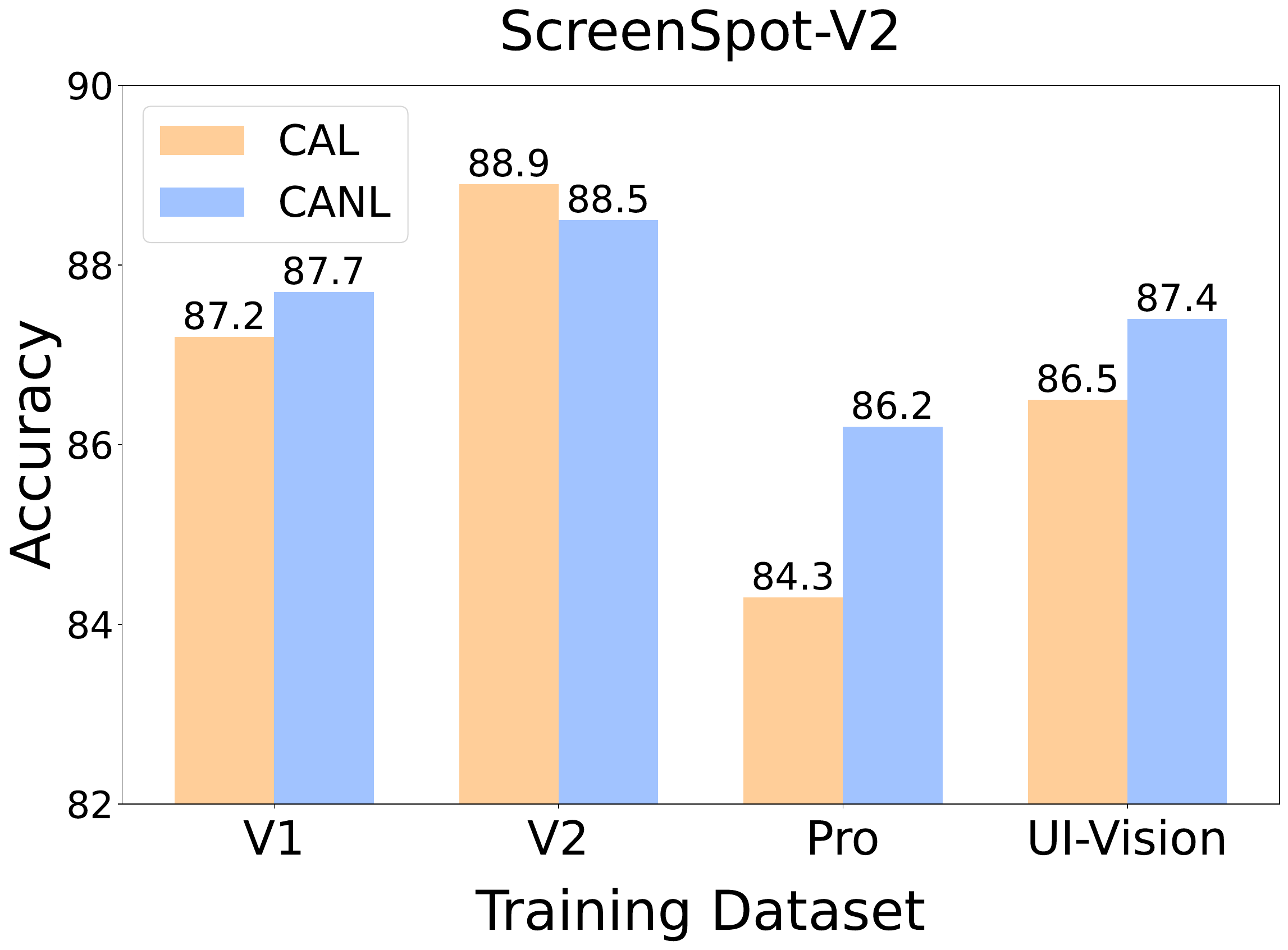}  %
        \label{fig:ood v2}
    \end{subfigure}
    \hfill  %
    \begin{subfigure}[b]{0.32\textwidth}
        \centering
        \includegraphics[width=\textwidth]{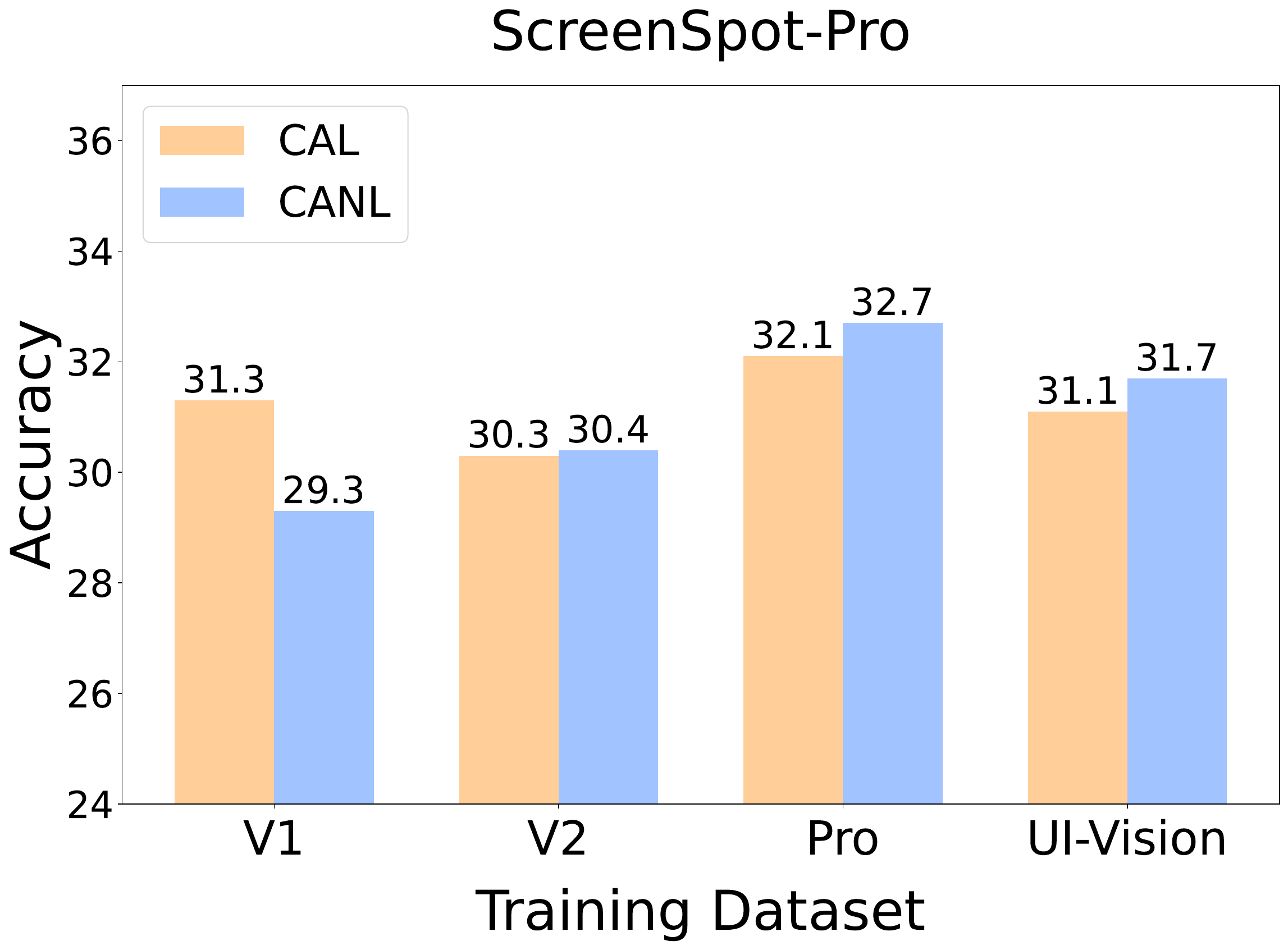}  %
        \label{fig:ood pro}
    \end{subfigure}
    \hfill
    \begin{subfigure}[b]{0.32\textwidth}
        \centering
        \includegraphics[width=\textwidth]{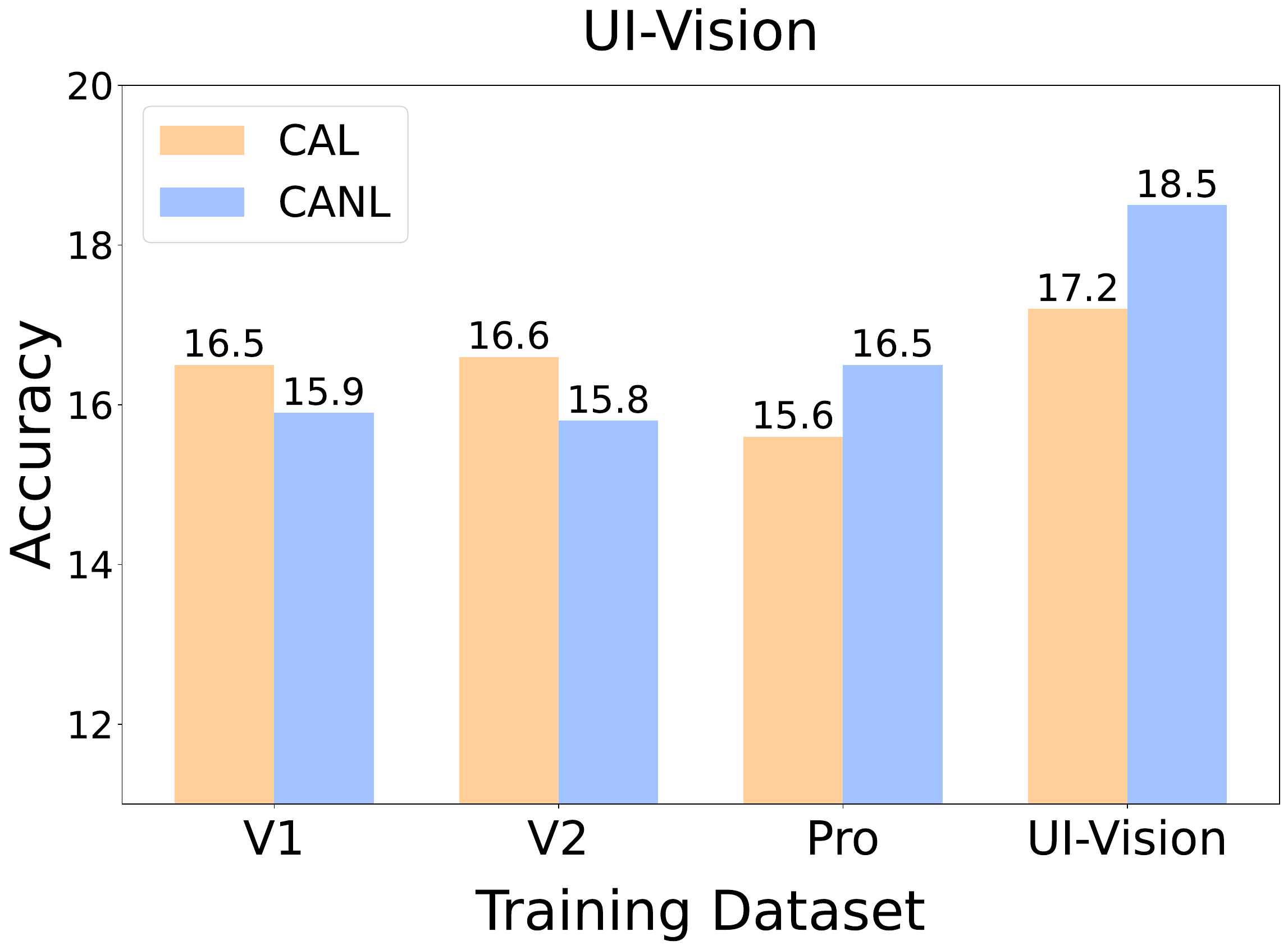}  %
        \label{fig:ood uiv}
    \end{subfigure}
    \caption{Cross-dataset generalization performance. Accuracy of our methods across different training sets when evaluated on ScreenSpot-V2, ScreenSpot-Pro, and UI-Vision.}  %
    \label{fig:ood}
\end{figure}
\noindent\textbf{Cross-dataset experiments demonstrate the generalization capability.} We train Qwen-2.5-VL-3B for one epoch on a source dataset and evaluate its transferability, as shown in \cref{fig:ood}; ScreenSpot-V1 is excluded due to documented annotation errors \cite{osatlas}. Training on UI-Vision yields gains of 5.3\% on ScreenSpot-V2 (87.4\% vs. 82.1\%) and 15.6\% on ScreenSpot-Pro (31.7\% vs. 16.1\%), while training on ScreenSpot-V2 improves UI-Vision by 4.6\%. These results confirm that our method avoids overfitting, achieving robust performance on unseen distributions. Notably, on more challenging benchmarks, CANL consistently outpaces CAL, highlighting its superior robustness in complex scenarios.

\noindent\textbf{Training on public datasets further validates generalization.} we extend our training to GroundCUA \cite{groundcua}, a publicly available benchmark. Specifically, we randomly sample 5k instances for a single epoch of training, with results summarized in \cref{tab:public train}. Our method exhibits robust generalization capability; even when trained on external public data, it yields substantial performance gains across evaluation benchmarks. This underscores that our label-free paradigm effectively captures the underlying spatial logic of GUI grounding rather than merely memorizing dataset-specific patterns.

\begin{table}[h]
\centering

\caption{Evaluation of our methods trained on publicly available datasets.}

\begin{tabular}{ccccc}
    \midrule
    \textbf{Method} & \textbf{ScreenSpot} & \textbf{ScreenSpot-V2} & \textbf{ScreenSpot-Pro} & \textbf{UI-Vision}\\
    \midrule
    Qwen-2.5-VL-3B & 77.6 & 82.1 & 16.1 & 12.0 \\
    \quad w/CAL & \textbf{84.6} & \textbf{87.3} & 32.6 & \textbf{16.7}\\
    \quad w/CANL & 83.5 & 86.7 & \textbf{33.4} & \textbf{16.7} \\
    \midrule
\end{tabular}
\label{tab:public train}
\end{table}

\begin{wraptable}{r}[-15pt]{0.35\textwidth}
\centering
\small
\setlength{\tabcolsep}{2pt} 

\caption{Success rate on AndroidWorld.}

\begin{tabular}{ccc}
    \toprule
    Planner & Grounding & SR\\ 
    \midrule
    \multirow{3}{*}{GPT-4o} & Qwen-2.5-VL-3B & 25.0 \\
    & \quad w/CAL & \textbf{30.2} \\
    & \quad w/CANL & 29.3 \\
    \bottomrule
\end{tabular}
\label{tab:android_world}
\end{wraptable}

\noindent\textbf{Online end-to-end evaluation demonstrates the effectiveness of our method in real-world scenarios.} We evaluate our approach on AndroidWorld \cite{androidworld}, a realistic platform comprising 116 tasks. To isolate the impact on GUI grounding, we adopt the SeeAct-V framework \cite{uground}, which decouples planning from precise coordinate prediction. Specifically, we evaluate the model previously trained on ScreenSpot-V2 \cite{osatlas} using our methods. As shown in \cref{tab:android_world}, our approaches yield a 4.3–5.2\% performance gain. CAL slightly outperforms CANL, since the simple training dataset produces more precise pseudo-labels. These results indicate that the enhanced grounding precision from our methods directly mitigates execution failures in multi-step sequences, demonstrating the practical efficacy of our methods for real-world GUI agents.

\section{Conclusion and Future Improvement}

This work introduces a label-free paradigm that successfully decouples GUI grounding from expensive manual labeling. By identifying coordinate-token confidence as a reliable proxy for accuracy, we proposed CAL and CANL to enable autonomous optimization. Our results—particularly the success of CANL—show that in sparse GUI environments, negative reinforcement can be more effective than error-prone positive pseudo-labels. However, an inherent limitation remains: while negative samples are highly reliable in challenging scenarios, the absence of accurately positive reward signal—compared to supervised training—may constrain the further scaling of model capabilities. Future research could focus on integrating more accurate positive feedback to achieve a more synergistic balance between avoiding errors and reinforcing optimal grounding behaviors.

\section*{Acknowledgement}

This work was supported by New Generation Artificial Intelligence-National Science and Technology Major Project (2025ZD0123100), National Natural Science Foundation of China (No. 62506332), "Pioneer" and "Leading Goose" R\&D Program of Zhejiang (NO. 2026C02A1223), and CCF-Ant-Research Fund.

\appendix
\section*{Ethics Statement}
This work focuses on advancing label-free reinforcement learning methods for GUI grounding tasks. Our research does not involve the collection or annotation of human subject data, nor does it utilize personally identifiable information. The datasets employed in this study are publicly available benchmarks, ensuring that no additional privacy or ethical risks are introduced. Potential misuse of our method, such as deploying GUI agents in malicious automation scenarios, should be carefully considered by practitioners. We encourage responsible application of our approach within research and development contexts that align with ethical guidelines and benefit broader society.

\section*{Reproducibility Statement}

To ensure reproducibility, we provide detailed descriptions of training configurations, hyperparameters, and evaluation protocols in the main paper and supplementary materials. All experiments are conducted on publicly available datasets. we report results averaged over multiple runs to account for variability. These steps are intended to facilitate faithful reproduction and fair comparison of our results.
\section{Appendix}

\subsection{Evaluation Details} Following \cite{ttrl}, we independently apply our methods on each benchmark to implement test-time reinforcement learning. On each dataset, we use all available original inputs (image and instruction) on each benchmark for label-free training. This section provides an overview of the benchmarks.
\begin{itemize}
    \item \textbf{ScreenSpot} \cite{seeclick} is a widely used benchmark for GUI grounding, which contains 1272 instructions across mobile, desktop and web domains.
    \item \textbf{ScreenSpot-V2} \cite{osatlas} is a enhanced version of ScreenSpot with error correction and re-annotation. It contains 1272 instructions across mobile, desktop and web domains.
    \item \textbf{ScreenSpot-Pro} \cite{screenspot-pro} is designed to rigorously evaluate the grounding capabilities of MLLMs in high-resolution professional settings. It contains 1581 samples, spanning 23 applications across five industries and three operating systems. 
    \item \textbf{UI-Vision} \cite{ui-vision} is a comprehensive, license-permissive benchmark for offline, fine-grained evaluation of computer use agents in real-world desktop environments. It provides three fine-to-coarse grained tasks: (1) element grounding; (2) layout grounding; and (3) action prediction. Since our work focuses on the model’s grounding capability for target elements, we use only the element grounding component during both training and evaluation. It introduce three grounding subtasks---basic, functional, and spatial---to assess different aspects of GUI understanding beyond simple textual queries. These three categories contain 1772, 1772, and 1935 instructions, respectively, totaling 5749 samples.
\end{itemize}

\subsection{Sparse Rewards vs. Dense Rewards Under Label-Free Setting}
\begin{figure}[htb]
    \centering
    \begin{subfigure}[b]{0.45\textwidth}
        \centering
        \includegraphics[width=\textwidth]{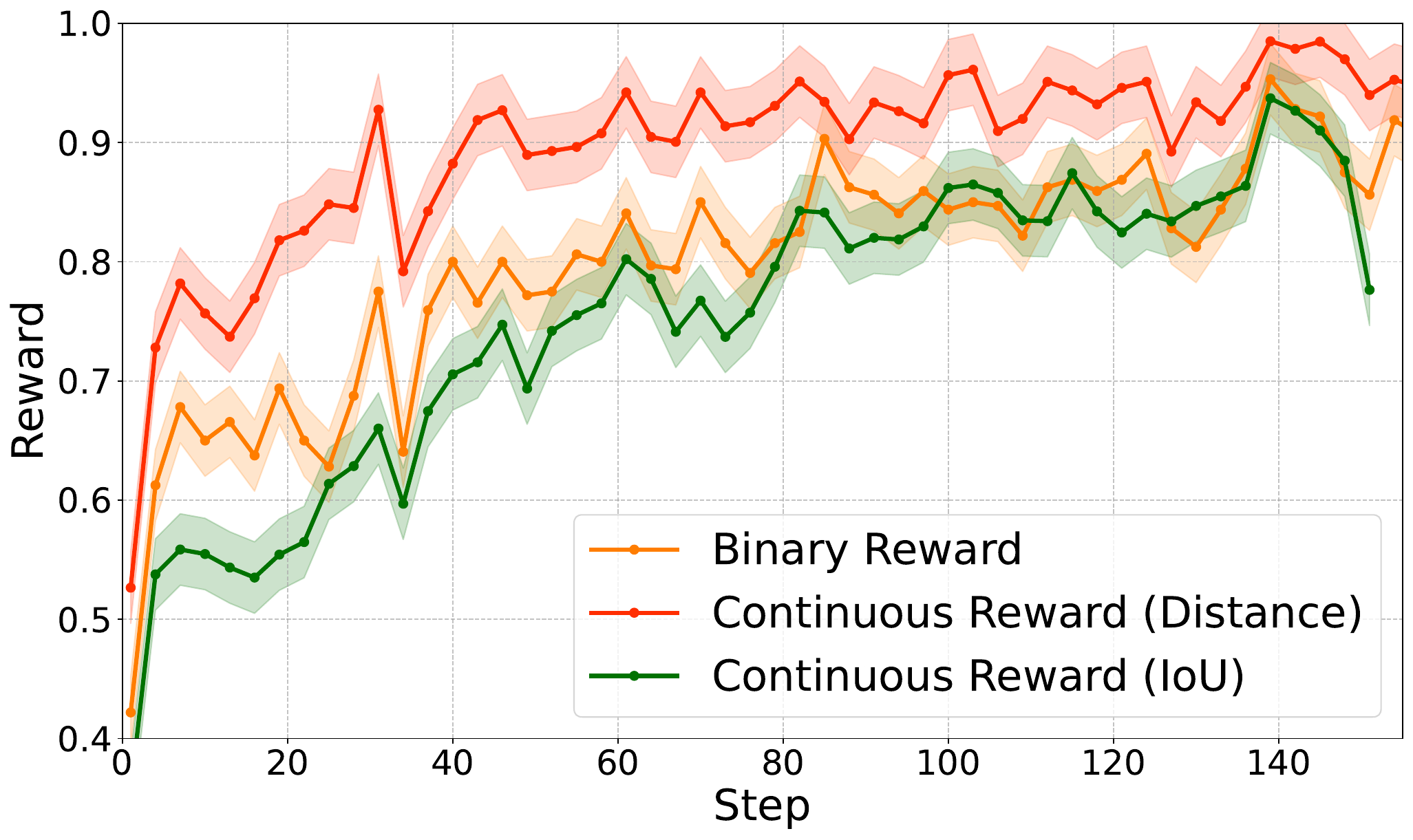}
        \label{fig:sparse and dense reward}
    \end{subfigure}
    \hfill  %
    \begin{subfigure}[b]{0.45\textwidth}
        \centering
        \includegraphics[width=\textwidth]{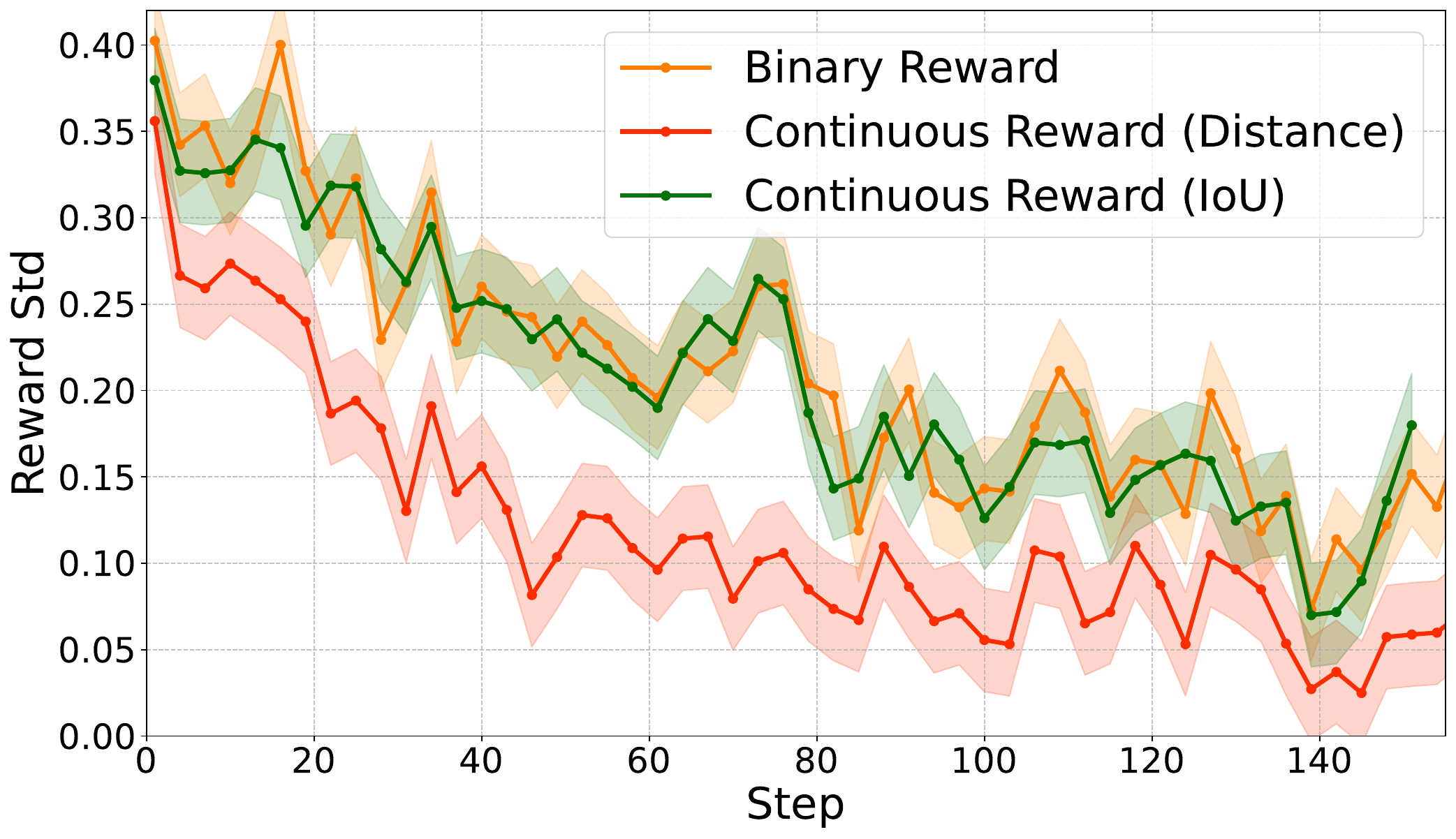}
        \label{fig:sparse and dense std}
    \end{subfigure}
    \caption{\textbf{Left}: Reward of binary reward and continuous reward during training on ScreenSpot-V2. \textbf{Right}: Reward std of binary reward and continuous reward during training on ScreenSpot-V2.}  %
    \label{fig:sparse and dense}
\end{figure}

In our proposed methods, the pseudo label is represented as a single point, making reward assignment based on point-to-point distance a natural choice, which corresponds to a dense reward scheme. In addition, all points can be dilated into a region according to the distance threshold $\tau$, and the elliptical IoU between each region and the pseudo-label region can be computed as a dense reward. Dense rewards provide fine-grained supervisory signals, thereby facilitating more effective model learning. As shown in \cref{fig:sparse and dense}, training with continuous rewards (Distance) accelerates convergence. However, as training progresses, its accuracy drops noticeably below that of binary rewards, as demonstrated in \cref{tab:dense acc}. Although the IoU-based reward scheme slows down the model’s convergence, the resulting performance gains are noticeably smaller compared to binary reward. We contend that under noisy labels, finer reward granularity increases susceptibility to error. Specifically, continuous fine-grained rewards encourage the model to align outputs closely with the pseudo label. Even when the pseudo label lies within the target bounding box, it rarely coincides with the true center of the bounding box, thereby introducing bias. More critically, when the pseudo label falls outside the bounding box, continuous rewards exacerbate the issue by further misleading the model.

\begin{table}[!htbp]
    \centering
    \caption{Accuracy of different reward type during training on ScreenSpot-V2.}
    \begin{tabular}{cccc}
    \midrule
    \textbf{Reward Type} & \textbf{50 Step} & \textbf{100 Step} & \textbf{159 Step} \\
    \midrule
    Binary Reward & 87.6 & 87.9 & 88.9 \\
    Continuous Reward (Distance) & 87.7 & 86.3 & 85.2 \\
    Continuous Reward (IoU) & 87.1 & 87.4 & 87.7 \\
    \midrule
\end{tabular}

    \label{tab:dense acc}
\end{table}

\subsection{Pure Positive Reinforcement Learning}
\begin{figure}[htb]
    \centering
    \begin{subfigure}[b]{0.45\textwidth}
        \centering
        \includegraphics[width=\textwidth]{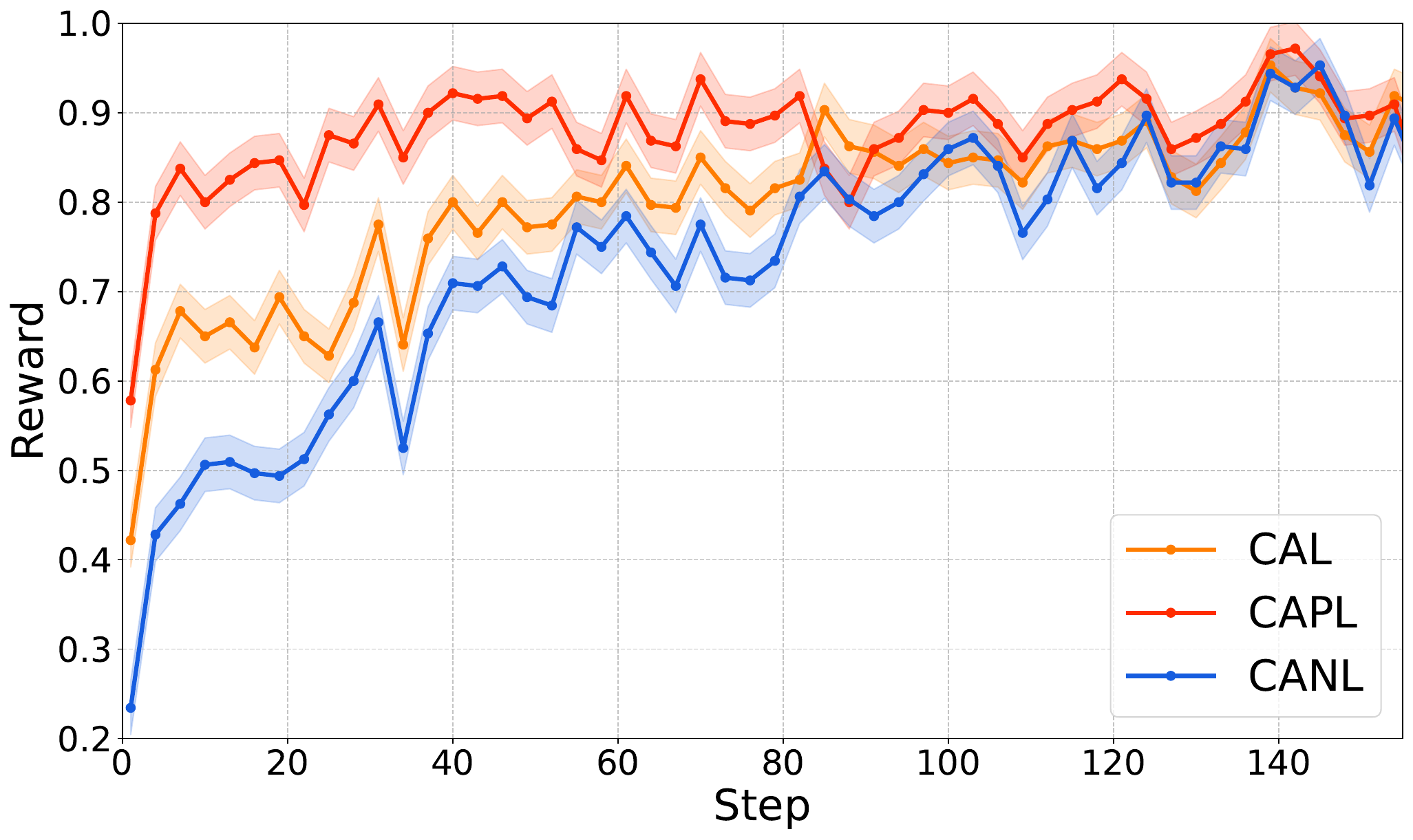}
        \label{fig:pure positive reward}
    \end{subfigure}
    \hfill  %
    \begin{subfigure}[b]{0.45\textwidth}
        \centering
        \includegraphics[width=\textwidth]{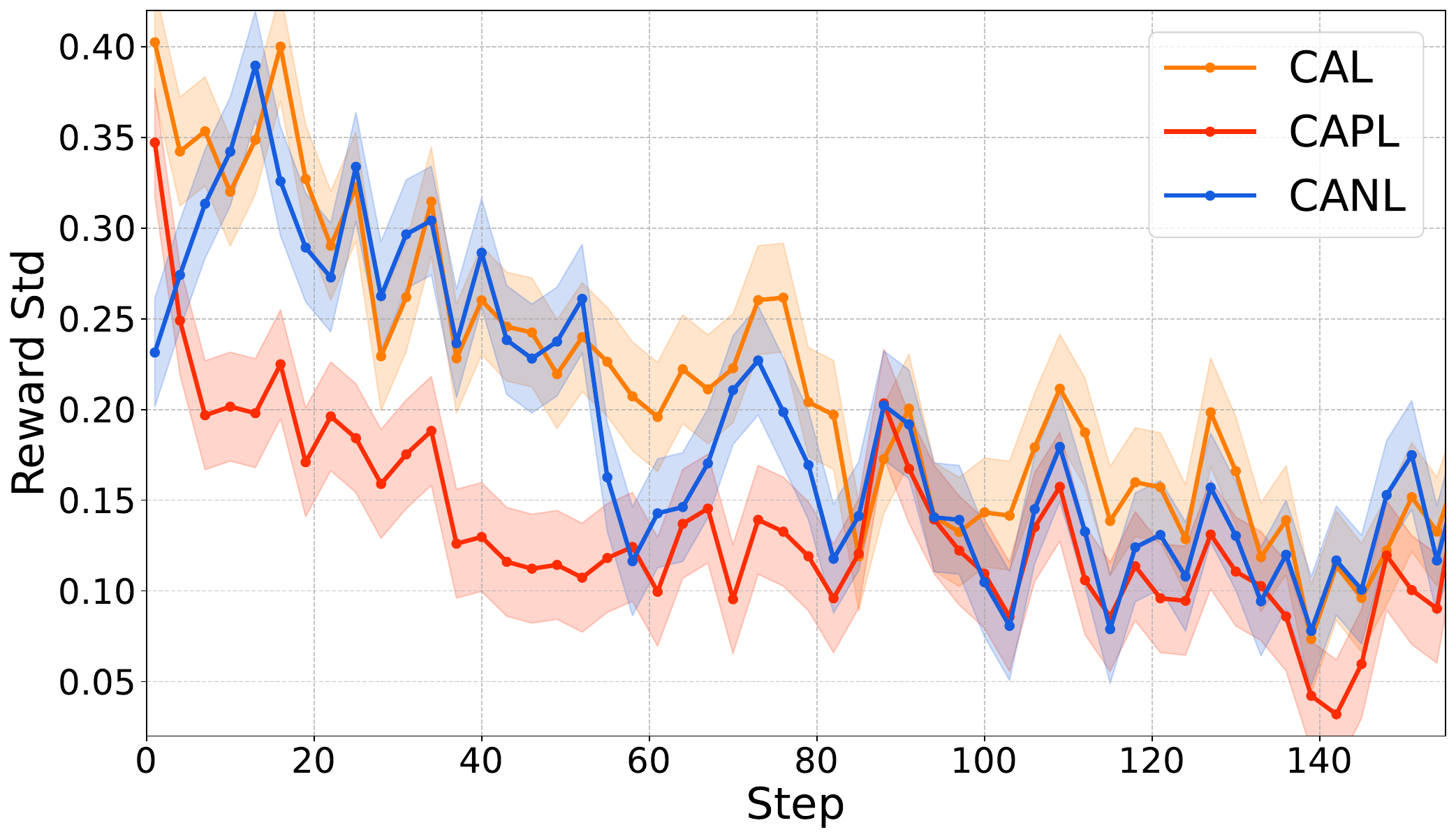}
        \label{fig:pure positive std}
    \end{subfigure}
    \caption{\textbf{Left}: Reward of CAL, CANL and pure positive learning during training on ScreenSpot-V2. \textbf{Right}: Reward std of CAL, CANL and pure positive learning during training on ScreenSpot-V2.}  %
    \label{fig:pure positive learning}
\end{figure}

We hypothesize that negative samples provide more reliable supervision than positive ones in label-free settings. To validate this, we conduct an ablation study using \textbf{C}onfidence-\textbf{A}nchored \textbf{P}ositive \textbf{L}earning (CAPL), which reinforces pseudo-labeled positive samples exclusively. As shown in \cref{fig:pure positive learning}, CAPL exhibits faster reward convergence; however, \Cref{tab:pure positive acc} reveals that accuracy gains remain marginal and eventually deteriorate as training progresses. This discrepancy suggests that rapid reward convergence merely reflects the model's tendency to collapse its predictions toward erroneous pseudo-labels, leading to severe overfitting to incorrect targets. Given the inherent noise in pseudo-labels, over-reliance on positive reinforcement misguides the optimization and triggers performance degradation. In contrast, while CANL converges more gradually, it significantly outperforms CAPL, confirming that negative signals offer more robust supervision when ground truth is absent. Notably, on the relatively simpler ScreenSpot-V2, positive samples retain a degree of reliability. Consequently, integrating both signals—as implemented in CAL—can further boost performance, striking a balance between cautious avoidance and constructive reinforcement.
\begin{table}[]
    \centering
    \caption{Accuracy of different sample selection mechanisms during training on ScreenSpot-V2.}
    \begin{tabular}{cccc}
    \midrule
    \textbf{Method} & \textbf{50 Step} & \textbf{100 Step} & \textbf{159 Step} \\
    \midrule
    CAPL & 86.3 & 87.2 & 86.7 \\
    CAL & 87.6 & 87.9 & 88.9 \\
    CANL & 88.4 & 88.0 & 88.5 \\
    \midrule
\end{tabular}

    \label{tab:pure positive acc}
\end{table}

\subsection{Reinforcement learning vs. Supervised Fine-Tuning.}

We generate pseudo-labels for Supervised Fine-Tuning (SFT) using the same sampling parameters as RL and apply identical training settings. Results, shown in \cref{tab:rl_sft}, reveal that SFT underperforms compared to RL, suggesting that RL better unlocks the model’s potential in pseudo-label settings. Unlike SFT’s fixed targets, GRPO uses relative advantage to optimize the policy, which mitigates the impact of noisy pseudo-labels, leading to better performance with imperfect data.

\begin{table}[]
    \centering
    \caption{Accuracy of RL and SFT on ScreenSpot-V2.} 
    \begin{tabular}{cc}
        \toprule
        Method & Accuracy \\ 
        \midrule
        Qwen-2.5-VL-3B & 82.1 \\
        \quad w/SFT & 84.6 \\
        \quad w/CAPL & 87.2 \\
        \quad w/CAL & \textbf{88.9} \\
        \quad w/CANL & \underline{88.5} \\
        \bottomrule
    \end{tabular}
\label{tab:rl_sft}
\end{table}

\subsection{Comparison of Different Pseudo-Label Construction Method}
\label{sec:pseudo_baseline_intro} 
\Cref{tab:tts_temperature} and \Cref{tab:tts_rollout} shows the performance of different pseudo-label construction methods with different temperature and rollout number. Coordinate-token confidence, Confidence represents selecting by coordinate-token confidence and averaged all-token confidence respectively. Coordinate-token Entropy and Entropy indicates selecting by coordinate-token entropy and averaged all-token entropy respectively. Random means selecting a predicted point randomly. Majority Voting selects as the answer the point that has the largest number of neighboring points within a predefined distance threshold. Following \cite{reguide}, we applied Kde, Center and Medoid. Kde selects the point with highest estimated density in the plane.  Center is simply getting the mean of all predicted coordinates. Medoid selects an actual predicted point that minimizes the sum of distances to all other predictions. As shown in \cref{tab:tts_temperature}, with the increase of temperature, the prediction points generated by the model become more inaccurate, leading to a decline in the performance of all methods. As the number of rollouts increases, the pseudo-labels constructed by each method become increasingly accurate, as shown in \cref{tab:tts_rollout}. Compared with other methods, our proposed approach achieves a substantial lead in performance under any temperature and number of rollouts, which demonstrated its effectiveness and robustness.

\begin{table}[t]
    \centering
    \caption{Performance of different pseudo-label construction methods with different temperature. Best and second-best results are shown in \textbf{Bold} and \underline{underline}, respectively.}
    \small
\begin{tabular}{cccccc}
    \midrule
    \textbf{Method} & \textbf{$T=0.6$} & \textbf{$T=0.7$} & \textbf{$T=0.8$} & \textbf{$T=0.9$} & \textbf{$T=1.0$} \\
    \midrule
    Coordinate-Token Confidence & \textbf{83.3} & \textbf{83.2} & \textbf{82.4} & \textbf{82.6} & \textbf{80.7} \\
    Confidence & 82.1 & 80.3 & 77.7 & 74.2 & 71.7 \\
    Coordinate-Token Entropy & 81.6 & 81.4 & \underline{80.9} & 78.2 & 77.4 \\
    Entropy & 82.1 & 81.6 & 80.3 & 77.7 & 76.6 \\
    KDE & 82.2 & \underline{82.1} & 79.7 & 80.0 & 78.2 \\
    Center & 74.5 & 70.8 & 68.2 & 62.0 & 55.0 \\
    Majority Voting & 81.6 & 80.5 & 80.0 & \underline{80.3} & \underline{79.6} \\
    Medoid & \underline{82.5} & 81.8 & 80.0 & 79.3 & 78.0 \\
    Random & 79.0 & 75.5 & 73.3 & 69.7 & 67.0 \\
    \midrule
\end{tabular}

    \label{tab:tts_temperature}
\end{table}

\begin{table}[t]
    \centering
    \caption{Performance of different pseudo-label construction methods with different rollout number $N$. Best and second-best results are shown in \textbf{Bold} and \underline{underline}, respectively.}
    \small
\begin{tabular}{ccccccc}
    \midrule
    \textbf{Method} & \textbf{$N=2$} & \textbf{$N=4$} & \textbf{$N=6$} & \textbf{$N=8$} & \textbf{$N=10$} & \textbf{$N=12$} \\
    \midrule
    Coordinate-Token Confidence & \textbf{76.7} & \textbf{81.5} & \textbf{82.8} & \textbf{83.2} & \textbf{83.3} & \textbf{83.9} \\
    Confidence & 74.8 & 78.4 & 80.0 & 80.3 & 78.9 & 78.6 \\
    Coordinate-Token Entropy & \underline{75.7} & 80.1 & 80.7 & 81.4 & 81.4 & 81.4 \\
    Entropy & 75.6 & 79.9 & 80.5 & 81.6 & 82.0 & 81.4 \\
    KDE & 75.6 & 80.1 & 81.4 & \underline{82.1} & 81.3 & 81.1 \\
    Center & 73.3 & 74.2 & 72.9 & 70.8 & 71.3 & 72.0 \\
    Majority Voting & 72.7 & 79.0 & 80.7 & 80.5 & 81.5 & 80.9 \\
    Medoid & 72.7 & \underline{80.6} & \underline{81.5} & 81.8 & \underline{82.5} & \underline{81.8} \\
    Random & 73.5 & 74.5 & 76.5 & 75.5 & 76.4 & 75.9\\
    \midrule
\end{tabular}

    \label{tab:tts_rollout}
\end{table}

\subsection{Comparison of Different Coordinate-Token Confidence Calculation}

\begin{wraptable}{r}[-15pt]{0.35\textwidth}
\centering
\small
\setlength{\tabcolsep}{4pt} 

\caption{Accuracy of different confidence calculation methods on ScreenSpot-V2.} 
\begin{tabular}{cc}
    \toprule
    Method & Accuracy \\ 
    \midrule
    Mean & \textbf{80.7} \\
    Product & \underline{80.4} \\
    Max & 76.4 \\
    Min & 78.1 \\
    \bottomrule
\end{tabular}
\label{tab:confidence_method}
\end{wraptable}

Prior work has proposed multiple approaches for computing confidence scores. Given the probability values of the generated coordinate tokens, we evaluate three confidence estimation methods: (1) Mean, which is the approach adopted in our training; (2) Product, which uses the N-th root of product of all coordinate-token probabilities as the confidence; (3) Max, which selects the highest probability among all coordinate-token probabilities; and (4) Min, which selects the lowest probability among all coordinate-token probabilities. We conduct experiments with temperature $T=1.0$ and rollout number $N=8$, and the results are presented in \cref{tab:confidence_method}. Experimental results indicate that directly computing the mean probability yields the most effective confidence measure. It strikes the best balance between penalizing uncertainty, providing the most reliable signal for pseudo-label selection.

\subsection{Downsampling Strategy}
\label{downsample}
During training, we sample 16 responses to construct pseudo-labels, followed by downsampling 8 to optimization, which is similar to \cite{ttrl}. It is a critical mechanism for overcoming the fundamental challenge of no ground-truth data in our label-free setting. Sampling 16 responses (rather than 8) is crucial for two reasons: it generates a sufficient variety of predictions, which is essential for discovering high-quality pseudo-labels and effective negative samples. Furthermore, downsampling 8 responses for optimization improves computational efficiency (in contrast to optimizing with all 16 responses) and enhances the negative signal quality for CANL. These 8 samples are not random, but are specifically selected as the farthest from the chosen pseudo-label. By restricting the policy update to these most distant samples, the negative reward signals are derived from the highest quality "incorrect" predictions, making the penalty mechanism highly effective for correcting errors.

\subsection{Training Hyperparameters}
\label{hyperparameters}
\begin{table}
\centering
    \small
    \caption{Training hyperparameters.}
    \begin{tabular}{lc}
    \midrule
    \textbf{Hyperparameter} & \textbf{Value} \\
    \midrule
    $\tau$ & 0.05 \\
    $\beta$ & 0.04 \\
    $T$ & 1.0 \\
    $top\_k$ & 50 \\
    $top\_p$ & 1.0 \\
    learning\_rate & 1e-6 \\
    bf16 & true \\
    torch\_dtype & bfloat16 \\
    data\_seed & 42 \\
    gradient\_checkpointing & true \\
    attn\_implementation & flash\_attention\_2 \\
    num\_train\_epochs & 1 \\
    max\_pixels & 12845056 \\
    \midrule
    \end{tabular}
    \label{tab:hyperparameters}
\end{table}

We provide detailed hyperparameter configurations to ensure reproducibility of our label-free training approach. \Cref{tab:hyperparameters} presents the core training parameters used across all experiments. The distance threshold $\tau$=0.05 was selected based on preliminary experiments to balance between capturing genuine positive samples and maintaining negative sample reliability. We employ a relatively conservative learning rate of 1e-6 with KL penalty $\beta$=0.04 to ensure stable policy updates during reinforcement learning. Sampling parameters ($T=1.0, top\_k=50, top\_p=1.0$) are configured to maximize response diversity, which proves critical for generating varied candidates for pseudo-label selection. To accommodate different dataset characteristics, we adjust gradient accumulation steps: 8 for ScreenSpot-V2, 4 for ScreenSpot-Pro, and 16 for UI-Vision, ensuring consistent effective batch sizes despite varying computational demands. All experiments utilize bfloat16 mixed precision training with gradient checkpointing and Flash Attention 2 for memory efficiency, enabling training on consumer-grade GPUs. The prompts are tailored to model capacity, with 3B models using a simpler single-point format while 7B models support multi-element detection, though both maintain the same coordinate-based output structure essential for our confidence computation.

\begin{tcolorbox}[
    title=3B Model Prompt,
    label=prompt_3b
]
\texttt{point to the instruction: \{Question\}, output its coordinates in JSON format \{\{"point\_2d": [x, y], "label": "object name/description"\}\}.}
\end{tcolorbox}

\begin{tcolorbox}[
    title=7B Model Prompt,
    label=prompt_7b
]
\texttt{Locate the UI element(s) for \{Question\}, output the coordinates using JSON format: [\{\{"point\_2d": [x, y]\}\}, ...]}
\end{tcolorbox}

\bibliographystyle{splncs04}
\bibliography{main}

\end{document}